\documentclass[lettersize,journal]{IEEEtran}
\usepackage{amsmath,amsfonts}
\usepackage{algorithmic}
\usepackage{algorithm}
\usepackage{array}
\usepackage[caption=false,font=normalsize,labelfont=sf,textfont=sf]{subfig}
\usepackage{textcomp}
\usepackage{stfloats}
\usepackage{url}
\usepackage{verbatim}
\usepackage{graphicx}
\usepackage{cite}
\usepackage{graphicx}
\usepackage{amsmath} 
\usepackage{graphicx}
\usepackage{multirow}  
\usepackage{colortbl}
\usepackage{threeparttable}
\usepackage{booktabs}
\usepackage{xcolor}
\usepackage{array}
\usepackage{arydshln}
\usepackage{mathrsfs}
\usepackage{amsfonts,amssymb,eucal}
\usepackage{graphics} 
\usepackage{epsfig} 
\usepackage{mathptmx} 
\usepackage{times} 
\usepackage{amsmath} 
\usepackage{amssymb}  
\usepackage{makecell}
\usepackage{tabularx}
\usepackage[colorlinks,
            linkcolor=blue,    
            citecolor=blue,    
            urlcolor=blue]    
            {hyperref}
\usepackage{cite}

\begin{document}

\title{RoGER-SLAM: A Robust Gaussian Splatting SLAM System for Noisy and Low-light Environment Resilience
}

\author{Huilin Yin, Zhaolin Yang, Linchuan Zhang, Gerhard Rigoll \textit{IEEE Fellow}, Johannes Betz

\IEEEcompsocitemizethanks{
  \IEEEcompsocthanksitem This work was supported by the National Natural Science Foundation of China under Grant No. 62433014 and No.62133011. 
  \IEEEcompsocthanksitem Huilin Yin and Zhaolin Yang are with the College of Electronic and Information Engineering, Tongji University, Shanghai 200092, China.
  \IEEEcompsocthanksitem Linchuan Zhang is with the Shanghai Research Institute for Intelligent Autonomous Systems, Tongji University, 201210 Shanghai, China.
  \IEEEcompsocthanksitem Gerhard Rigoll is with the Chair of Human-Machine Communication, Technical University of Munich, Munich 80333, Germany.
  \IEEEcompsocthanksitem Johannes Betz is with the Professorship of Autonomous Vehicle Systems, Technical University of Munich, Garching 85748, Germany, Munich Institute of Robotics and Machine Intelligence (MIRMI).
}
}



\markboth{IEEE Transactions on Instrumentation and Measurement}%
{Shell \MakeLowercase{\textit{et al.}}: A Sample Article Using IEEEtran.cls for IEEE Journals}


\maketitle

\begin{abstract}

The reliability of Simultaneous Localization and Mapping (SLAM) is severely constrained in environments where visual inputs suffer from noise and low illumination. Although recent 3D Gaussian Splatting (3DGS) based SLAM frameworks achieve high-fidelity mapping under clean conditions, they remain vulnerable to compounded degradations that degrade mapping and tracking performance. A key observation underlying our work is that the original 3DGS rendering pipeline inherently behaves as an implicit low-pass filter, attenuating high-frequency noise but also risking over-smoothing. Building on this insight, we propose RoGER-SLAM, a robust 3DGS SLAM system tailored for noise and low-light resilience. The framework integrates three innovations: a Structure-Preserving Robust Fusion (SP-RoFusion) mechanism that couples rendered appearance, depth, and edge cues; an adaptive tracking objective with residual balancing regularization; and a Contrastive Language-Image Pretraining (CLIP)-based enhancement module, selectively activated under compounded degradations to restore semantic and structural fidelity. Comprehensive experiments on Replica, TUM, and real-world sequences show that RoGER-SLAM consistently improves trajectory accuracy and reconstruction quality compared with other 3DGS-SLAM systems, especially under adverse imaging conditions.
\end{abstract}

\begin{IEEEkeywords}
3D Gaussian Splatting (3DGS), 3D reconstruction, Robustness in degraded environments, Simultaneous Localization and Mapping (SLAM).
\end{IEEEkeywords}

\section{Introduction}
\IEEEPARstart{S}{imultaneous} 
Localization and Mapping (SLAM) persists as a foundational capability for autonomous systems operating in unstructured environments, with mission-critical applications in robotics, augmented reality, and autonomous driving.  
With the advancement of SLAM research, the focus has gradually shifted beyond localization accuracy toward achieving photorealistic and structurally consistent map reconstruction.
In this context, neural rendering techniques such as Neural Radiance Fields (NeRF)~\cite{mildenhall2020nerf} have demonstrated impressive photorealistic reconstruction capabilities by representing scenes as continuous volumetric fields.  
However, NeRF-based SLAM approaches~\cite{wang2021nerfslam},~\cite{xu2022nice},~\cite {zhu2023coslam},~\cite {zheng2023point},~\cite{murez2020atlas},~\cite{timKN-SLAM} often suffer from high computational cost, slow convergence and weak structural regularization, which limit their applicability in real-time and degraded scenarios.

To overcome these limitations, 3D Gaussian Splatting (3DGS)~\cite{kerbl20233dgs} has emerged as an efficient alternative, achieving real-time, high-fidelity scene reconstruction.  
3DGS excels at modeling fine-scale geometry and photometric details with real-time differentiable rendering, but its pipeline lacks explicit structural constraints and multiview consistency.  
As a result, 3DGS-based SLAM~\cite{chen2023gsslam},~\cite{zimmermann2024monogs},~\cite{hsieh2023splatam},~\cite{timLVIGS} remains highly sensitive to perceptual degradations such as sensor noise and low-light conditions, which can corrupt high-frequency appearance and destabilize tracking. Since such degradations frequently occur in real-world scenarios, achieving robust SLAM under these adverse conditions remains particularly critical for safety-sensitive applications~\cite{engel2018dso,zhang2023lowlightslam},~\cite{dpvins}.

\begin{figure}
    \centering
    \includegraphics[width=0.95\linewidth]{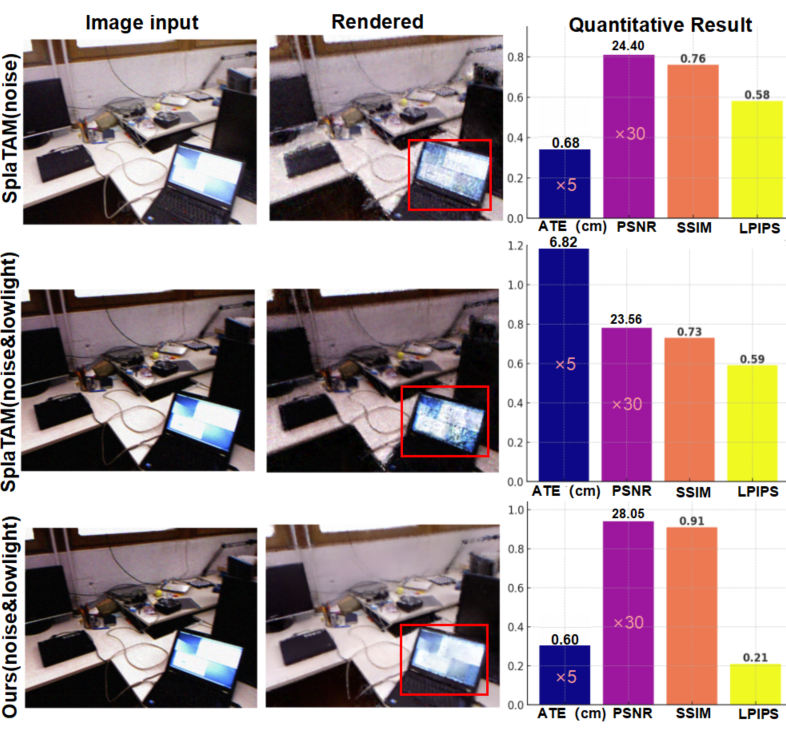}
    \caption {Qualitative and quantitative comparison under noise and low-light conditions. The first column shows the degraded input images, the second column presents the rendered reconstructions, and the third column reports the quantitative results. Compared with SplaTAM under noise inputs (top) and compound inputs (middle), our method (bottom) achieves significantly sharper reconstructions with clearer structural details and superior quantitative performance in terms of ATE, PSNR, SSIM, and LPIPS.}
    \label{fig:1}
\end{figure}

This paper proposes a robust 3DGS-based SLAM system explicitly designed to withstand noisy and low-light conditions.
Our approach integrates three core innovations: a structure-preserving robust fusion mechanism, an adaptive tracking objective, and a selectively triggered Contrastive Language-Image Pretraining
(CLIP)-based enhancement module.
To the best of our knowledge, this work also presents the first systematic quantitative study of how compounded noise and low-light jointly affect the accuracy and stability of 3DGS-based SLAM.

A key observation underlying our work is that the 3DGS rendering pipeline inherently exhibits a low-pass filtering effect. In this pipeline, each Gaussian contributes to the final image through an opacity-weighted accumulation along the viewing ray. The alpha compositing of multiple overlapping Gaussians therefore acts as a weighted integration process, analogous to Gaussian smoothing in classical image processing. This property provides a valuable perspective for improving robustness, as it naturally attenuates high-frequency noise while preserving coarse scene structures. However, because this effect is implicit and non-structural, it can over-smooth fine geometric details and remains insufficient under compounded degradations. To overcome this limitation, we propose the Structure-Preserving Robust Fusion mechanism, which constructs a structure-aware pseudo-supervision view by combining rendered images, normalized depth, and gradient-based edges within a pyramid framework. This fused representation explicitly reinforces geometric fidelity while suppressing corruption, thereby enhancing robustness under adverse conditions.

Nevertheless, under extreme degradations such as severe noise or very low illumination, the structure-aware pseudo-supervision view alone remains insufficient. To further enhance robustness, we introduce a CLIP-based enhancement module that is conditionally triggered, ensuring high-level feature restoration only when adverse conditions are detected.

In parallel, we design an adaptive tracking method that dynamically balances color and depth residuals with regularization, preventing degenerate weight assignments and improving trajectory stability across diverse photometric and geometric conditions.

To summarize, our contributions are as follows: 
\begin{itemize}



\item A fusion mechanism is proposed that integrates rendered images, depth maps, and edge cues into a structure-aware pseudo-supervisory signal. This design preserves geometric fidelity and suppresses high-frequency corruption, thereby improving mapping robustness under degraded conditions.

\item A residual balancing objective with regularization is formulated to dynamically adjust the weighting between color and depth residuals. This strategy prevents degenerate weight assignments and ensures stable camera pose estimation across diverse photometric and geometric conditions.

\item A selectively triggered enhancement module based on CLIP is developed to restore high-level semantic features when severe noise or low illumination is detected. This selective activation avoids redundant computation in normal scenarios, striking a balance between robustness and efficiency.

\item To the best of our knowledge, the first systematic quantitative study of the coupled impact of noise and low light on 3DGS-based SLAM is presented. The analysis highlights the unique challenges posed by compounded degradations and validates the robustness of the proposed framework. Relative to the baseline, the method yields a 50\% improvement in tracking performance on clean Replica data and a 91\% improvement under noisy and low-light conditions.
\end{itemize}

Qualitative and quantitative results are shown in Fig~\ref{fig:1}. The remainder of this paper is organized as follows. Sec.~\ref{sec:2} reviews related work on dense SLAM and robustness under adverse conditions. Sec.~\ref{sec:3} presents the proposed RoGER-SLAM framework in detail. Sec.~\ref{sec:4} reports experimental results on both synthetic and real-world datasets, followed by ablation studies. Sec.~\ref{sec:5} concludes the paper with discussions and future directions.

\section{related work}\label{sec:2}
\subsection{Visual Dense SLAM}
A central challenge in SLAM research lies in reconstructing geometrically consistent and photometrically dense 3D environments from sparse 2D visual input. Early dense SLAM systems adopt explicit geometry such as TSDF voxels~\cite{voxels} and surfels~\cite{surfels} to realize real-time mapping and tracking from RGB-D streams. KinectFusion~\cite{kinectfusion2011} introduces volumetric fusion with GPU acceleration, enabling accurate indoor reconstruction, while ElasticFusion~\cite{elasticfusion} extends this paradigm with surfel fusion and non-rigid map refinement for improved robustness. Despite their effectiveness in geometric reconstruction, such methods remain constrained in photorealistic rendering and novel-view synthesis beyond the captured viewpoints.

Neural implicit methods subsequently bring photorealistic view synthesis into SLAM. NICE-SLAM~\cite{zhu2022nice} proposes hierarchical neural encoding with geometric priors for scalable dense SLAM; Co-SLAM~\cite{zhu2023coslam} combines coordinate and sparse-parametric encodings with global bundle adjustment for real-time RGB-D SLAM; Point-SLAM~\cite{zheng2023point} uses a neural point-cloud parameterization for dense mapping. Despite impressive reconstructions, these methods incur high computational cost and can be brittle under photometric degradations.

Fortunately, 3DGS~\cite{kerbl20233dgs} enables real-time differentiable rendering and high-fidelity novel view synthesis with explicit, anisotropic Gaussians. Building on this representation, several SLAM systems have been developed. SplaTAM\cite{hsieh2023splatam} combines splatting with tracking and mapping with differentiable rendering-based optimization; GS-SLAM~\cite{chen2023gsslam} integrates 3DGS into a dense SLAM pipeline with loop closure; MonoGS~\cite{zimmermann2024monogs} adapts Gaussian splatting to monocular SLAM. These methods demonstrate strong performance in clean environments, but they predominantly rely on photometric residuals and lack explicit mechanisms to enforce structural consistency, which makes them sensitive to degradations.

Recently, efforts have been made to extend 3DGS-SLAM towards robustness under real-world disturbances. DG-SLAM~\cite{xu2024dgslam} incorporates motion-aware Gaussian management and hybrid pose optimization to improve stability in dynamic environments, while WildGS-SLAM~\cite{zheng2025wildgs} introduces uncertainty-guided tracking to suppress the influence of moving objects under monocular input. Although these methods improve robustness, they primarily target scene dynamics rather than photometric degradations such as sensor noise and low-light addressed in this work.

\subsection{Interference Suppression in Visual SLAM}
Visual SLAM performance significantly degrades in the presence of noise and low-light, which frequently co-occur in real-world scenarios. Recent studies have explored the integration of image enhancement techniques into SLAM frameworks to improve robustness. Twilight-SLAM~\cite{zhang2023twilightslam} integrates low-light enhancement networks into SLAM pipelines; ULL-SLAM~\cite{lu2023ullslam} designs an end-to-end enhancement front-end for underwater low-light environments; and Low-Light SLAM~\cite{liu2022lowlightslam} couples learned enhancement with feature matching for robustness across illumination changes. However, these enhancement-based SLAM systems remain tailored to conventional pipelines and are not directly applicable to 3DGS-SLAM, where optimization is tightly coupled with rendering consistency rather than pre-processed image quality.

\begin{figure*}
    \centering
    \includegraphics[width=1\linewidth]{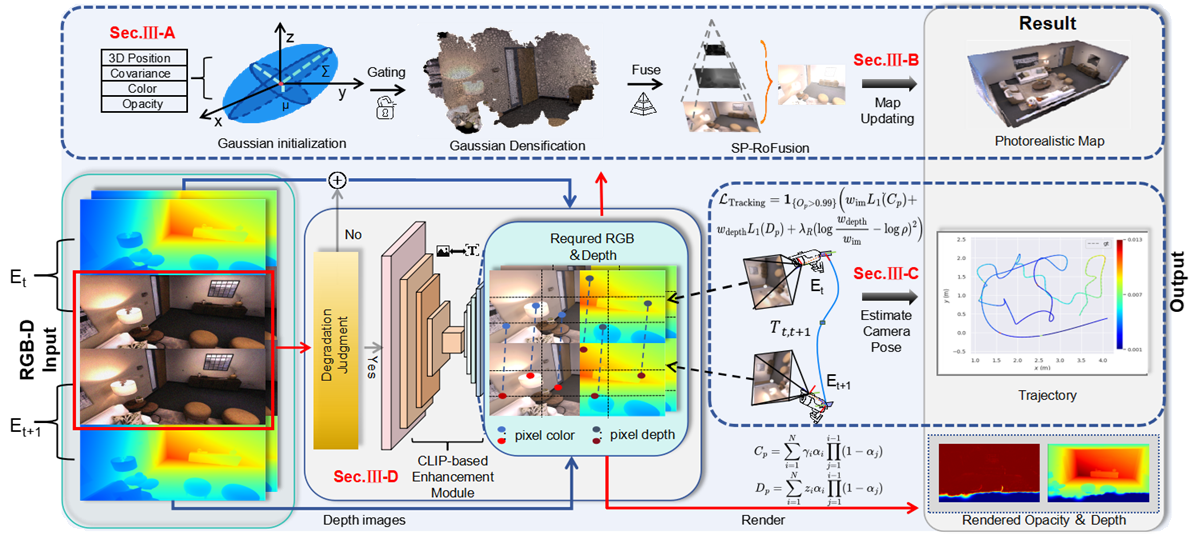}
    \caption{Framework Overview.
Given RGB-D inputs,
Gaussian spheres are initialized and densified (Sec.III-A), and updated through structure-preserving robust fusion (Sec.III-B). 
Camera poses are estimated via an adaptive tracking objective (Sec.III-C). A degradation judgment module selectively activates the CLIP-based enhancement module (Sec.III-D). 
The pipeline produces photorealistic reconstructions and consistent trajectories.}
    \label{fig:main}
\end{figure*}

From the image enhancement perspective, widely used methods such as EnlightenGAN~\cite{jiang2021enlightengan}, Zero-DCE~\cite{guo2020zerodce}, and its improved variant Zero-DCE++~\cite{guo2023zerodce++} are capable of correcting exposure without requiring paired supervision. However, these methods primarily optimize perceptual quality criteria and do not enforce the geometric consistency that is essential for SLAM backend.
On the noise side, classical denoising algorithms such as DnCNN~\cite{zhang2017dncnn} and FFD-net~\cite{zhang2018ffdnet} have shown strong performance under synthetic Gaussian noise. Nevertheless, these methods are typically trained under simplified assumptions and often generalize poorly to the photon-limited conditions encountered in low-light SLAM scenarios. More recent research has focused on physically grounded noise modeling. Raw-domain analyses have established that sensor noise in low illumination is well described by a Poisson–Gaussian process~\cite{foi2008noise}, where photon shot noise dominates as illumination decreases and read noise contributes as an additional constant component. NODE~\cite{ma2019node} exploit this decomposition for extreme low-light denoising. Despite these advances, such photometric characteristics remain largely unexplored in the rendering-based optimization pipelines of 3DGS-SLAM.

Moreover, public benchmarks seldom provide datasets that combine sensor noise and low-light while offering ground-truth camera trajectories. ExDark~\cite{loh2019exdark}, for example, characterizes low-light image distributions for recognition tasks but cannot be used for SLAM evaluation. Consequently, the compounded effects of noise and low-light on rendering fidelity and tracking robustness in 3DGS-based SLAM have not been systematically investigated. This absence provides the central motivation for the robust framework introduced in this paper.

\section{method}\label{sec:3}

Our SLAM framework builds entirely upon the 3DGS representation, integrating four pivotal modules: 
a differentiable Gaussian rendering backend, a structure-preserving robust fusion module, 
an adaptive tracking objective, and a dual-condition-triggered CLIP-based enhancement module. 
As illustrated in Fig.~\ref{fig:main}, these modules are integrated into a unified pipeline that enables robust and accurate 
reconstruction and trajectory estimation, even under severe noise and low-illumination conditions.
\subsection{3D Gaussian Representation}
To reconstruct a photorealistic map, a collection of 3D Gaussian points are used to model the local geometry and appearance of the environment. Each 3D Gaussian is defined as an ellipsoid characterized by its mean position ($\mu$), covariance matrix ($\Sigma$), opacity ($\alpha$), and color. The density function for a single Gaussian $i$ is expressed as:
\begin{equation}
G_i(x) = \alpha_i \exp\left(-\frac{1}{2} (x - \mu_i)^T \Sigma_i^{-1} (x - \mu_i)\right),
\end{equation}
where $x \in \mathbb{R}^3$ represents a point in 3D space. 

 \textit{Initialization}: The system generates Gaussian points frame-by-frame from input RGB-D images. Each 3D Gaussian is projected to 2D image coordinates using camera intrinsics $K$ and extrinsics $P$. The center position $\mu_i$ and projected covariance $\Sigma'_i$ account for perspective distortion:
\begin{equation}
\mu_i = K^{-1} \begin{bmatrix} x_i & y_i & 1 \end{bmatrix}^T \cdot d_i \, \, , \quad
\Sigma'_i = J P \Sigma_i P^T J^T \, ,
\end{equation}
where $J$ is the Jacobian of the perspective projection, $d_i$ represents the depth of the pixel $(x_i,y_i)$. Projected Gaussians are sorted by depth and alpha-blended to compute pixel colors ($C_p$) and depths ($D_p$):
\begin{equation}
C_p = \sum_{i=1}^{N}\gamma_i \alpha_i \prod_{j=1}^{i-1}(1 - \alpha_j), \quad
D_p = \sum_{i=1}^{N} z_i \alpha_i \prod_{j=1}^{i-1}(1 - \alpha_j),
\end{equation}
where $\gamma_i$ is the color of the $i$-th Gaussian and $N$ denotes depth-ordered Gaussians, $z_i$ is the depth of $\mu_i$ in camera coordinates. 

\textit{Densification}: To maintain map completeness, new Gaussians are dynamically added to previously unmodeled regions during camera motion. Specifically, we detect such regions by evaluating the rendered opacity $O_p$:
\begin{equation}
O_p = \sum_{i=1}^{N} \alpha_i \prod_{j=1}^{i-1} (1 - \alpha_j).
\end{equation}

The opacity term reflects the visibility contribution of each Gaussian. If a region has an opacity value below 0.5 and its predicted depth deviates from the ground-truth by more than 50 times the median error, new Gaussians are inserted to recover the missing structure.

To avoid redundant insertions from uninformative viewpoints, we propose a multi-scale gating mechanism applied prior to densification. For each candidate frame, opacity maps are rendered at three spatial resolutions $s \in \{1.0, 0.5, 0.25\}$. At each scale, an accumulated per-pixel opacity map $O_p^{(s)}$ is computed. The multi-scale importance score is then obtained through weighted fusion across all scales:

\begin{equation}
\text{IMP}_{\text{fuse}} = \sum_{s} w^{(s)} \cdot \frac{1}{|P|} \sum_{p \in P} O_p^{(s)},
\end{equation}
where $P$ denotes all valid pixels, and typical weights are $w^{(1.0)} = 0.5$, $w^{(0.5)} = 0.3$, and $w^{(0.25)} = 0.2$. Only when $\text{IMP}_{\text{fuse}} $ over 0.01 does the frame trigger densification, during which new Gaussians are placed via opacity-guided spatial sampling.

\begin{figure}
    \centering    \includegraphics[width=1\linewidth]{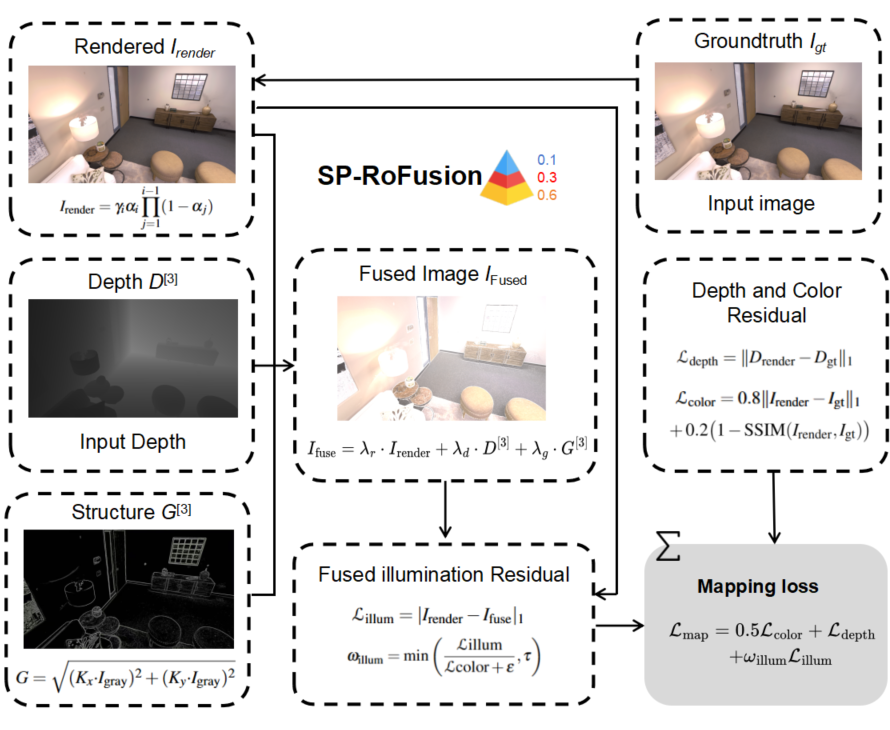}
    \caption{Structure-preserving robust fusion (SP-RoFusion) module. 
The rendered image $I_{\text{render}}$, input depth $D^{[3]}$, and structural edge map $G^{[3]}$ (where $[3]$ denotes channel replication to three dimensions) are linearly fused into $I_{\text{fuse}}$ with weights $\alpha$, $\beta$, and $\gamma$. 
Depth, color, and illumination residuals are then computed, and combined into the final mapping loss $\mathcal{L}_{\text{map}}$ with an adaptive illumination weight.
}
    \label{fig:3}
\end{figure}

\subsection{Structure-Preserving Robust Fusion and Mapping}

We propose a Structure-Preserving Robust Fusion strategy (SP-RoFusion), which enhances mapping stability under perceptual degradation. Specifically, we construct a pseudo-supervised target image $I_{\text{fuse}}$ by integrating rendered appearance, geometric depth, and structural edge cues. The overall pipeline of SP-RoFusion including the fusion formulation and loss design is illustrated in  Fig.~\ref{fig:3}. This fused representation serves as a robust supervisory signal that retains essential geometric information while suppressing photometric distortions.

Standard 3D Gaussian Splatting frameworks focus on photometric fidelity for optimization, often neglecting fine-grained structural details, especially under noise or low-light conditions. However, structural cues such as edges and geometric discontinuities play a crucial role in preserving scene integrity and guiding accurate geometry reconstruction. By incorporating structure-aware supervision via edge-enhanced pseudo targets, our method enforces spatial regularization and improves convergence stability in challenging areas.


Let $I_{\text{render}} \in \mathbb{R}^{3 \times H \times W}$ denote the rendered RGB image, $D_{\text{gt}} \in \mathbb{R}^{1 \times H \times W}$ denote the ground-truth depth map, and $G \in \mathbb{R}^{1 \times H \times W}$ denote the structural edge map. We construct the fused pseudo image $I_{\text{fuse}}$ as follows:
\begin{equation}
I_{\text{fuse}} = \lambda_r \cdot I_{\text{render}} 
+ \lambda_d \cdot \textit{}{Norm}(\text{Rep}_3(D_{\text{gt}})) 
+ \lambda_g \cdot \textit{}{Norm}(\text{Rep}_3(G)),
\end{equation}
where $\lambda_r$, $\lambda_d$, and $\lambda_g$ are fusion weights, $\text{Rep}_3(\cdot)$ denotes 1-to-3 channel replication, and $\textit{}{Norm}(\cdot)$ represents min-max normalization to [0,1].

The edge map $G$ is computed using the Sobel operator on the grayscale projection of the rendered image:
\begin{equation}
G = \sqrt{(K_x \text{·} I_{\text{gray}})^2 + (K_y \text{·} I_{\text{gray}})^2},
\end{equation}
where $K_x$ and $K_y$ are standard Sobel kernels along the $x$ and $y$ axes, and $I_{\text{gray}}$ is obtained by averaging $I_{\text{render}}$ across RGB channels.

To enforce robust photometric supervision, we define the following losses:
\begin{align}
\mathcal{L}_{\text{illum}} &= \| I_{\text{render}} - I_{\text{fuse}} \|_1, \,\,\,\,
\mathcal{L}_{\text{depth}} = \| D_{\text{render}} - D_{\text{gt}} \|_1, \\
\mathcal{L}_{\text{color}} &= 0.8 \| I_{\text{render}} - I_{\text{gt}} \|_1
+ 0.2 \big( 1 - \text{SSIM}(I_{\text{render}}, I_{\text{gt}}) \big), \\
\mathcal{L}_{\text{map}} &= 0.5 \mathcal{L}_{\text{color}} 
+ \mathcal{L}_{\text{depth}} 
+ \omega_{\text{illum}} \mathcal{L}_{\text{illum}},
\end{align}
where ${D}_{\text{render}}$ is the rendered depth, SSIM is the structural similarity index between two values.
The dynamic illumination weight $\omega_{\text{illum}}$ is computed as
\begin{equation}
\omega_{\text{illum}} = \min \left( \frac{\mathcal{L}_{\text{illum}}}{\mathcal{L}_{\text{color}} + \epsilon}, \tau \right),
\end{equation}
where $\epsilon$ avoids division by zero and $\tau$ limits the weight to prevent it from dominating the mapping loss. This structure-aware supervision strategy preserves crucial geometric cues, thereby significantly enhancing the robustness and fidelity of the SLAM system.

\subsection{Adaptive Camera Tracking Objective}  
Camera tracking aims to estimate the camera’s pose at the time of capturing the current input RGB-D image. The camera pose is initialized using a constant velocity forward projection strategy:  
\begin{align}
E_{t+2} = E_{t+1} + (E_{t+1} - E_{t}) \,\, ,
\end{align}
where \(E_t\) denotes the camera pose at frame \(t\). This leverages temporal continuity to provide robust initial estimates.  

\begin{figure*}
    \centering
    \includegraphics[width=1\linewidth]{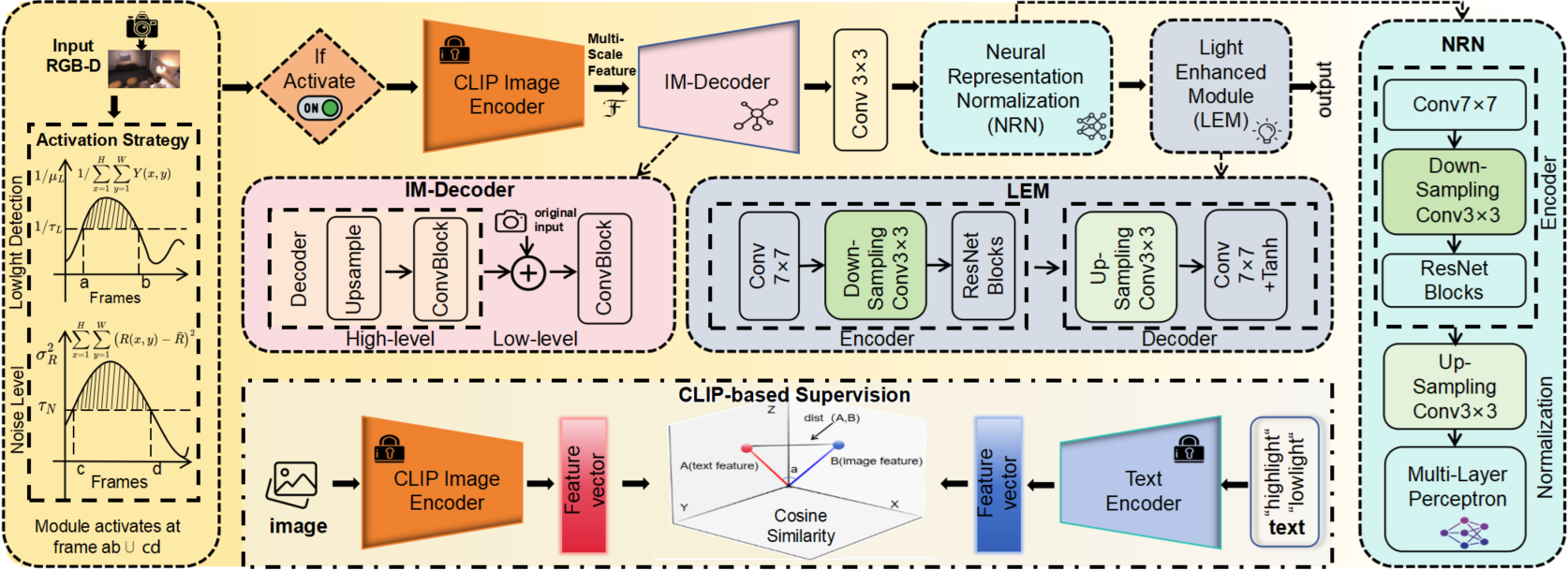}
    \caption{ Workflow of the CLIP Image Enhancement Module. 
The module addresses both noise reduction and low-light enhancement, and is selectively activated through a dual-condition judgment strategy based on global brightness and residual noise variance. 
A CLIP-based image encoder extracts multi-scale features from degraded inputs, which are reconstructed by the image decoder to suppress noise. 
The Neural Representation Normalization (NRN) further mitigates degradation artifacts, while the Light Enhancement Module (LEM) restores brightness and structural details. 
The enhancement process is constrained by CLIP-based supervision, aligning visual and text features via cosine similarity. 
Note that for visualization, the light-condition judgment is presented in reciprocal form for clarity, while in our implementation it is applied directly on the raw brightness values.
}
    \label{fig:4}
\end{figure*}

We adopt a dense direct tracking strategy that minimizes color and depth residuals to optimize the camera pose. To ensure reliability, only pixels with high opacity are used for supervision.
In 3DGS-based methods, fixed residual weights are commonly used, but a static color-to-depth weighting is not guaranteed to generalize across diverse scenes due to variations in texture, depth range, and noise conditions. Therefore, a dynamic reweighting mechanism is proposed as follows:
\begin{equation}
w_{\text{im}} = \frac{\gamma_{\text{im}}}{\mathcal{L}_{\text{color}} + \gamma_{\text{im}}}, \quad 
w_{\text{depth}} = \frac{\gamma_{\text{depth}}}{\mathcal{L}_{\text{depth}} + \gamma_{\text{depth}}},
\end{equation}
where 
 $\gamma_{\text{im}}$
  and 
 $\gamma_{\text{depth}}$
  are residual sensitivity coefficients that modulate the sensitivity of the respective weights to residual magnitudes. 

 To prevent these weights from diverging to extreme values, we implement a regularization-based adjustment mechanism. Specifically, prior experience guides the weighting of image residuals, ensuring that the ratio of residual weights remains within a reasonable range. To maintain mapping quality and reduce computation time, the system fixes Gaussian parameters and focuses on optimizing the tracking-related loss:
\begin{equation}
\begin{split}
\mathcal L_{\text{Tracking}} &= \mathbf{1}_{\{O_p > 0.99\}} \Big( 
w_{\text{im}} L_1(C_p) +
w_{\text{depth}} L_1(D_p)  
\\
&\quad  
+\lambda_R( 
\log \frac{w_{\text{depth}}}{w_{\text{im}}}  
- 
\log\rho 
)^2 \Big),
\end{split}
\end{equation}
where \(\mathbf{1}_{\{\cdot\}}\) is the indicator function, \(\rho\) is the pre-calibrated ideal ratio. This adaptive weighting strategy enables the system to robustly adapt to varying illumination and texture richness across scenes, ensuring stable convergence and better pose accuracy.

\subsection{CLIP Image Enhancement Module}

In practice, the coupled presence of strong noise and low-light severely undermines the effectiveness of the proposed SP-RoFusion module, leading to unstable supervision and degraded reconstructions. To address this challenge, we further introduce a CLIP-based image enhancement module, as illustrated in Fig.~\ref{fig:4}, including a denoising network with a frozen CLIP encoder and a multi-layer decoder, and a low-light enhancement module supervised by multi-modal text guidance. Leveraging the unique properties of CLIP, this module achieves both denoising and low-light enhancement, and is highly adaptable to diverse complex environments.

The denoising model primarily consists of a frozen CLIP image encoder and a multi-level learnable image decoder. Recent work reveals that
CLIP’s frozen encoder features exhibit distortion-invariant and content-related properties~\cite{clipdenoising2024}. The frozen encoder extracts noise-robust multi-scale features from the input noisy image. The decoder is composed of upsampling and convolutional layers. Multi-scale features are concatenated with decoder features to merge multi-scale information. Since the noisy input image contains rich image details, it is added to the decoder as an additional image feature. Finally, the denoised image \(I_d\) is obtained by performing convolutional operations on the processed features. The loss for training is the $L_1$ loss between denoised image \(I_d\) and clean image \(I_c\):
\begin{align}
\mathcal L _\text{denoising}=  L_1(I_d,I_c) \,.
\end{align}

The low-light enhancement network consists of a neural implicit normalization preprocessing step and an encoder-decoder enhancement module, supervised by a CLIP pretrained model from a text-guided perspective. 
Inspired by~\cite{Nerco}, the low-light enhancement network incorporates Neural Representation Normalization (NRN) to predict degradation to a uniform level before enhancement. NRN ensures adaptability to varying degradation levels, enabling the model to handle images under diverse low-light conditions. We treat each pixel $p$ together with its spatial coordinates $C$ and feature embeddings $E$ as the input to a decoding function 
$\mathcal F_{\text{MLP}}$. Formally, we define the reconstructed image $I_{\text{NR}}$ as:
\begin{align}
I_{\text{NR}}(p) = \mathcal F_{\text{MLP}}(C_p, E_p) \,.
\end{align}

Additionally, semantic prior supervision is utilized from a pretrained vision-language model. It consists of a text encoder and an image encoder, two prompts are manually designed: one for low-light images and the other for high-light images. The text and image encoders extract feature vectors of the same size from the prompts and images, and the cosine similarity between these vectors is computed to measure their differences. The training encourages the image vectors to be similar to the prompt \textit{high-light image} and dissimilar to the prompt \textit{low-light image}. The final loss function for light enhancement consists of three components:
\begin{align}
\mathcal L_{\text{LE}}  = L_1(I_{\text{NR}} , I_L)+L_1(I_H , I_L)+ \mathcal L_{\text{CLIP}}\, ,
\end{align}
\begin{align}
\mathcal L_{\text{CLIP}} = \frac{\langle \text{Enc}(I_L), \text{Enc}(T_H) \rangle}{||\text{Enc}(I_L)|| \,||\text{Enc}(T_H)||} - \frac{\langle \text{Enc}(I_L), \text{Enc}(T_L) \rangle}{||\text{Enc}(I_L)|| \, ||\text{Enc}(T_L)||}\, ,
\end{align}
where \(I_L\) is the input low-light image, \(I_H\) is the output enhanced image, \(T_H\) is the prompt \textit{high-light image},  \(T_L\) is the prompt \textit{low-light image}, and Enc is the encoding operation to extract feature vectors.

\textbf{Selective activation.} Our system adopts a selective enhancement strategy to balance robustness and efficiency. In normal scenarios with either no noise or only mild, real sensor noise, 
our SP-RoFusion pipeline can already reconstruct reliable geometry and produce visually satisfactory results.
However, in the presence of complex synthetic noise or extremely low-light conditions, 
the base pipeline may degrade due to the dominance of unstructured high-frequency noise
or insufficient brightness cues. 
To selectively activate the enhancement module, 
we adopt a dual-condition trigger combining global brightness~\cite{loh2019exdark} 
and residual noise variance~\cite{liu2012noise}.

First, the global brightness $\mu_L$ is computed as the mean grayscale intensity:
\begin{equation}
    \mu_L = \frac{1}{HW}\sum_{x=1}^{H}\sum_{y=1}^{W} Y(x,y),
\end{equation}
where $Y(x,y)$ denotes the grayscale value of pixel $(x,y)$. 
Low-light conditions are indicated when $\mu_L < \tau_L$.
We empirically set the low-light threshold $\tau_L$ to 80, aligning with prior literature that characterizes brightness levels below 90 as low-light or visually degraded conditions~\cite{loh2019exdark, lee2022unsupervised, xu2020learning}. This choice also provides a margin to avoid triggering enhancement under too mild darkening, ensuring that CLIP Enhancement is only invoked under really poor illumination.

Second, the residual variance $\sigma^2_R$ is estimated to approximate the unstructured noise level:
\begin{equation}
    \hat{Y}(x,y) = \textit{}{median\_filter}\big(Y(x,y), 3\times 3\big),
\end{equation}
\begin{equation}
    R(x,y) = Y(x,y) - \hat{Y}(x,y),
\end{equation}
\begin{equation}
    \sigma^2_R = \frac{1}{HW}\sum_{x=1}^{H}\sum_{y=1}^{W}\big(R(x,y)-\bar R\big)^2,
\end{equation}
where the \textit{median\_filter} denotes a $3\times3$ median filter applied to the input image $Y(x,y)$, $\bar R$ is the mean residual intensity.
This non-reference estimator follows the principle of blind noise estimation~\cite{liu2012noise} and aligns with the noise distribution observed in real datasets. According to the DND dataset~\cite{dnd2017}, 
real sensor noise in natural images mostly exhibits a standard deviation of 
$15$–$25$ in 8-bit intensity, 
rarely exceeding $25$ even under high ISO settings.  
When measured via residual variance, 
this corresponds to $\sigma^2_R < 30$ in our estimation.  
Therefore, we empirically set $\tau_N = 30$ to separate natural sensor noise 
from synthetic or extreme degradations, 
ensuring that the enhancement module is triggered only for frames 
with unstructured high-variance noise.

Finally, the CLIP enhancement module is triggered if:
\begin{equation}
    \mu_L < \tau_L \quad \text{or} \quad \sigma^2_R > \tau_N,
\end{equation}
with $\tau_L = 80$ and $\tau_N = 30$ in our implementation. 
This design ensures that CLIP enhancement is only invoked for extremely low-light 
or high-variance synthetic noise inputs, 
while normal or mildly noisy frames rely solely on SP-RoFusion for efficient processing.

\begin{figure*}
    \centering
    \includegraphics[width=0.95\linewidth]{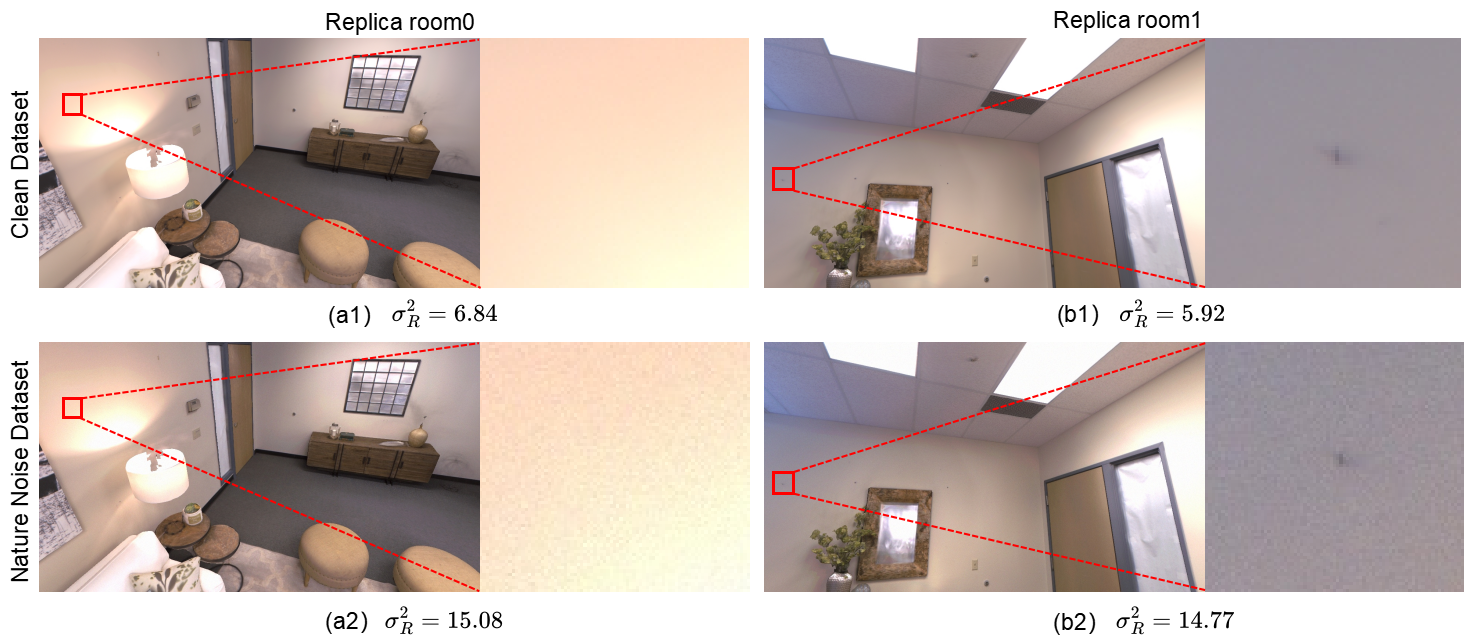}
    \caption{Examples from the constructed natural-noise dataset. 
columns (a) and (b) show two representative scenes, where the top row (a1, b1) corresponds to clean inputs and the bottom row (a2, b2) contains noisy counterparts generated with shot and read noise. 
The residual variance $\sigma_R^2$ quantifies the noise level, demonstrating the increase from mild to stronger perturbations after noise synthesis.
}

    \label{fig:5}
\end{figure*}

\begin{figure*}
    \centering
    \includegraphics[width=0.95\linewidth]{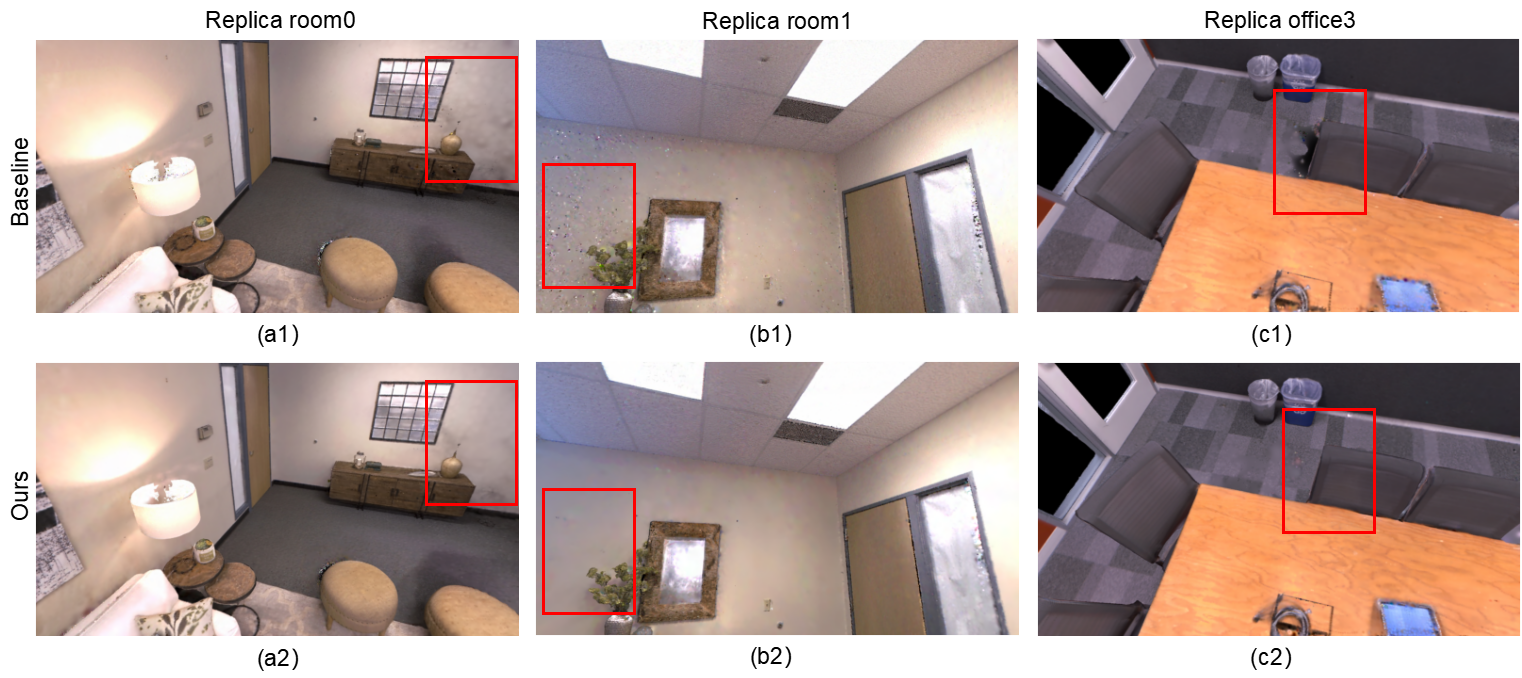}
    \caption{Qualitative comparison under real sensor noise. 
The top row (a1–c1) shows results from SplaTAM, while the bottom row (a2–c2) presents results from our RoGER-SLAM on the same scenes: (a) room0, (b) room1, and (c) office3. 
Red boxes highlight regions where SplaTAM exhibits noticeable artifacts or structure distortions, whereas our method produces cleaner geometry and reduced noise interference.}

    \label{fig:6}
\end{figure*}

\begin{figure*}
    \centering
    \includegraphics[width=0.95\linewidth]{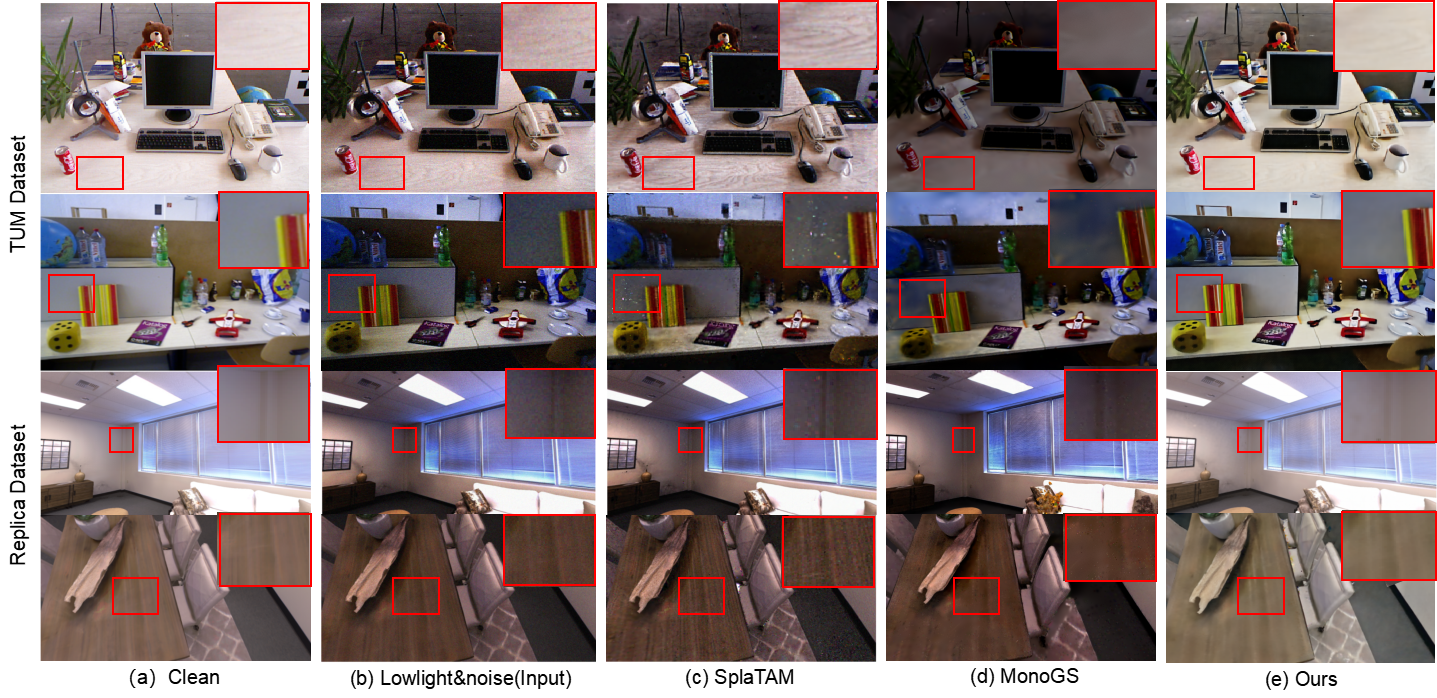}
    \caption{Reconstruction Performance on Replica and TUM Datasets.
(a) Clean dataset images; (b) Noise and low-light inputs; Subsequent columns: Reconstructed results. Compared to baseline methods, our system achieves higher rendering quality and photorealistic illumination aligned with human visual perception.}
    \label{fig:7}
\end{figure*}

\section{Experiments}\label{sec:4}

\subsection{Experimental Setup}
\textbf{Hardware and Implementation.}
All experiments are conducted on a workstation equipped with an Intel i9-13900 CPU, 32 GB RAM, and an NVIDIA RTX 4090 GPU. Our system is implemented in PyTorch with CUDA acceleration.

\textbf{Datasets.}
We evaluate our method on both synthetic and real-world benchmarks. The Replica\cite{Replica} dataset and the TUM-RGBD\cite{Tum-RGBD} dataset are used as standard clean benchmarks. To systematically investigate robustness under perceptual degradations, we construct three additional variants:
(1) a \textit{Replica-natural-noise} dataset with mild sensor-like noise to emulate realistic degradation;
(2) a \textit{Replica-low-light-noise} dataset with compounded strong noise and low illumination;
(3) a \textit{TUM-low-light-noise} dataset with similar compounded degradations.
These datasets allow us to progressively evaluate robustness under clean, noisy, and severely degraded conditions.

\textbf{Evaluation Metrics.}
We report results using both tracking accuracy and mapping fidelity metrics. Camera pose accuracy is measured by Absolute Trajectory Error (ATE). Mapping quality is quantified by Peak Signal-to-Noise Ratio (PSNR), Structural Similarity Index (SSIM), and Learned Perceptual Image Patch Similarity (LPIPS).

\begin{table}[t]
\large
\renewcommand\arraystretch{1}
\centering
\resizebox{1\linewidth}{!}{
\begin{threeparttable}  
\renewcommand{\arraystretch}{1.3} 
    \caption{\large Camera Tracking Result on Clean Replica (ATE RMSE $\downarrow$ [cm]). 
    Lower is better. 
    \textcolor{green!50!black}{\textbf{Green}}=best, 
    \textcolor{yellow!50!black}{Yellow}=second, 
    \textcolor{orange!80!black}{Orange}=third (R0–R2 and Of0–Of4 correspond to room0–2 and office0–4 sequences of Replica)}
    \label{tab:1}
    \begin{tabular}{lcccccccccc}
    \cmidrule[0.8pt]{1-11}
    & Method & Avg. & R0 & R1 & R2 & Of0 & Of1 & Of2 & Of3 & Of4 \\
    \cmidrule[0.8pt]{1-11}
    \multirow{4}{*}{\makecell{NeRF \\ based}} 
    & NICE SLAM  & 1.06 & 0.97 & 1.31 & 1.07 & 0.88 & 1.00 & 1.06 & 1.10 & 1.13 \\
    & Co-SLAM    & 0.99 & 0.77 & 1.04 & 1.09 & 0.58 & 0.53 & 2.05 & 1.49 & 0.84 \\
    & Point SLAM & 0.54 & 0.56 & 0.47 & \cellcolor{orange!30}0.30 & \cellcolor{green!30}\textbf{0.35} & 0.62 & 0.55 & 0.72 & 0.73 \\
    & Vox Fusion & 0.54 & \cellcolor{orange!30}0.40 & 0.54 & 0.54 & 0.50 & 0.46 & 0.75 & 0.50 & \cellcolor{orange!30}0.60 \\
    \cmidrule[0.4pt]{1-11}
    \multirow{4}{*}{\makecell{3DGS \\ based}} 
    & GS-SLAM    & \cellcolor{orange!30}0.50 & 0.48 & 0.53 & 0.33 & 0.52 & \cellcolor{orange!30}0.41 & 0.59 & 0.46 & 0.70 \\
    & MonoGS     & 0.58 & 0.44 & \cellcolor{yellow!30}0.32 & \cellcolor{yellow!30}0.31 & 0.44 & 0.52 & \cellcolor{green!30}\textbf{0.23} & \cellcolor{yellow!30}0.17 & 2.25 \\
    & SplaTAM    & \cellcolor{yellow!30}0.36 & \cellcolor{yellow!30}0.31 & \cellcolor{orange!30}0.40 & \cellcolor{yellow!30}0.29 & 0.47 & \cellcolor{yellow!30}0.27 & \cellcolor{orange!30}0.29 & \cellcolor{orange!30}0.32 & \cellcolor{yellow!30}0.55 \\
    & Ours       & \cellcolor{green!30}\textbf{0.24} & \cellcolor{green!30}\textbf{0.23} & \cellcolor{green!30}\textbf{0.21} & \cellcolor{green!30}\textbf{0.21} & \cellcolor{yellow!30}0.37 & \cellcolor{green!30}\textbf{0.20} & \cellcolor{yellow!30}0.24 & \cellcolor{green!30}\textbf{0.15} & \cellcolor{green!30}\textbf{0.35} \\
    \cmidrule[0.8pt]{1-11}
    \end{tabular}
\end{threeparttable}}
\end{table}

{\scriptsize
\begin{table}[t]
\centering
\resizebox{\linewidth}{!}{
\begin{threeparttable}  
    \caption{Camera Tracking Result on Clean TUM (ATE RMSE $\downarrow$ [cm]). 
    \textcolor{green!50!black}{\textbf{Green}}=best, 
    \textcolor{yellow!50!black}{Yellow}=second, 
    \textcolor{orange!80!black}{Orange}=third. “-” indicates missing results due to unavailable settings in the baselines 
    } 
    \label{tab:2}
    \begin{tabular}{lccccccc}
    \cmidrule[0.4pt]{1-8}
    & Method & Avg. & \makecell{fr1/ \\ desk} & \makecell{fr1/ \\ desk2} & \makecell{fr1/ \\ room} & \makecell{fr2/ \\ xyz} & \makecell{fr3/ \\ off} \\
    \cmidrule[0.4pt]{1-8}
    \multirow{4}{*}{\makecell{NeRF \\ based}} 
    & NICE SLAM   & 15.87 & 4.26 & 4.99 & 34.49 & 31.73 & 3.87 \\
    & Co-SLAM     & \cellcolor{orange!30}8.38  & \cellcolor{yellow!30}2.70 & \cellcolor{yellow!30}4.57 & 30.16 & 1.90 & \cellcolor{yellow!30}2.60 \\
    & Point SLAM  & 8.92  & 4.34 & \cellcolor{green!30}\textbf{4.54} & 30.92 & 1.31 & \cellcolor{orange!30}3.48 \\
    & Vox Fusion  & 11.31 & 3.52 & 6.00 & \cellcolor{orange!30}19.53 & 1.49 & 26.01 \\
    \cmidrule[0.4pt]{1-8}
    \multirow{4}{*}{\makecell{3DGS \\ based}} 
    & GS-SLAM     & --    & 3.30 & --    & --    & \cellcolor{orange!30}1.30 & 6.60 \\
    & MonoGS     & -- & \cellcolor{green!30}\textbf{1.50} & -- & -- & 1.44 & \cellcolor{green!30}\textbf{1.49} \\
    & SplaTAM     & \cellcolor{yellow!30}5.48  & 3.35 & 6.54 & \cellcolor{yellow!30}11.13 & \cellcolor{yellow!30}1.24 & 5.16 \\
    & Ours        & \cellcolor{green!30}\textbf{4.79} & \cellcolor{yellow!30}2.70 & \cellcolor{orange!30}4.83 & \cellcolor{green!30}\textbf{10.89} & \cellcolor{green!30}\textbf{1.21} & 4.34 \\
    \cmidrule[0.4pt]{1-8}
    \end{tabular}
\end{threeparttable}}
\end{table}}

\begin{table*}[t]
\centering
\caption{Quantitative Results under Clean and Natural Noise Conditions.Baseline is SplaTAM~\cite{hsieh2023splatam}, Diff. denotes the difference $\text{noise} - \text{clean}$. R0–R2 and Of0–Of4 correspond to room0–2 and office0–4 sequences of Replica (ATE RMSE $\downarrow$ [cm])}
\label{tab:3}
\setlength{\tabcolsep}{3pt}
\renewcommand{\arraystretch}{1.05}
\resizebox{\textwidth}{!}{
\begin{tabular}{ll
ccc ccc ccc ccc ccc ccc ccc ccc}
\toprule
\multicolumn{2}{c}{} 
& \multicolumn{3}{c}{R0} 
& \multicolumn{3}{c}{R1} 
& \multicolumn{3}{c}{R2} 
& \multicolumn{3}{c}{Of0} 
& \multicolumn{3}{c}{Of1} 
& \multicolumn{3}{c}{Of2} 
& \multicolumn{3}{c}{Of3} 
& \multicolumn{3}{c}{Of4} \\
\cmidrule(lr){3-5}\cmidrule(lr){6-8}\cmidrule(lr){9-11}
\cmidrule(lr){12-14}\cmidrule(lr){15-17}\cmidrule(lr){18-20} 
\cmidrule(lr){21-23}\cmidrule(lr){24-26}
& & clean & noise & Diff. 
  & clean & noise & Diff. 
  & clean & noise & Diff. 
  & clean & noise & Diff. 
  & clean & noise & Diff. 
  & clean & noise & Diff. 
  & clean & noise & Diff. 
  & clean & noise & Diff. \\
\midrule
\multirow{4}{*}{\textbf{Baseline}}
& ATE/cm$\downarrow$  
& 0.33 & 0.62 & 0.29
& 0.29 & 0.40 & 0.11
& 0.29 & 0.30 & \textbf{0.01}
& 0.49 & 0.53 & \textbf{0.04}
& 0.24 & 0.27 & 0.03
& 0.29 & 0.35 & 0.06
& 0.31 & 0.38 & \textbf{0.07}
& 0.55 & 0.59 & 0.04 \\
& PSNR$\uparrow$  
& 31.21 & 27.37 & -3.84
& \textbf{33.61} & 31.85 & -1.76
& \textbf{34.85} & \textbf{34.27} & \textbf{-0.58}
& \textbf{38.35} & 36.11 & -2.24
& \textbf{38.90} & 37.28 & -1.62
& \textbf{32.09} & 30.72 & \textbf{-1.37}
& 30.34 & 28.88 & -1.46
& 31.75 & \textbf{30.48} & \textbf{-1.27} \\
& SSIM$\uparrow$  
& \textbf{0.98} & 0.93 & -0.05
& \textbf{0.96} & \textbf{0.94} & -0.02
& 0.97 & 0.95 & -0.02
& 0.97 & 0.95 & -0.02
& 0.97 & \textbf{0.97} & 0.00
& \textbf{0.96} & \textbf{0.94} & -0.02
& 0.95 & 0.93 & -0.02
& \textbf{0.94} & \textbf{0.93} & -0.01 \\
& LPIPS$\downarrow$ 
& 0.09 & 0.25 & 0.16
& \textbf{0.10} & \textbf{0.30} & \textbf{0.20}
& 0.09 & 0.38 & 0.30
& 0.09 & 0.26 & \textbf{0.17}
& 0.10 & 0.19 & \textbf{0.09}
& \textbf{0.10} & 0.28 & 0.18
& \textbf{0.11} & 0.26 & 0.15
& \textbf{0.15} & 0.34 & 0.19 \\
\midrule
\multirow{4}{*}{\textbf{Ours}}
& ATE/cm$\downarrow$  
& \textbf{0.23} & \textbf{0.23} & \textbf{0.00}
& \textbf{0.21} & \textbf{0.25} & \textbf{0.04}
& \textbf{0.21} & \textbf{0.23} & 0.02
& \textbf{0.37} & \textbf{0.41} & \textbf{0.04}
& \textbf{0.20} & \textbf{0.21} & \textbf{0.01}
& \textbf{0.24} & \textbf{0.26} & \textbf{0.02}
& \textbf{0.15} & \textbf{0.24} & 0.09
& \textbf{0.35} & \textbf{0.38} & \textbf{0.03} \\
& PSNR$\uparrow$  
& \textbf{31.93} & \textbf{29.52} & \textbf{-2.41}
& 32.51 & \textbf{31.08} & \textbf{-1.43}
& 34.63 & 32.96 & -1.67
& 38.14 & \textbf{36.11} & \textbf{-2.03}
& 38.42 & \textbf{37.12} & \textbf{-1.30}
& 31.31 & 29.44 & -1.87
& \textbf{30.69} & \textbf{29.54} & \textbf{-1.15}
& \textbf{32.01} & 30.12 & -1.89 \\
& SSIM$\uparrow$  
& 0.97 & \textbf{0.96} & \textbf{-0.01}
& \textbf{0.96} & \textbf{0.94} & \textbf{-0.02}
& \textbf{0.98} & \textbf{0.97} & \textbf{-0.01}
& \textbf{0.98} & \textbf{0.96} & \textbf{-0.02}
& \textbf{0.98} & \textbf{0.97} & \textbf{-0.01}
& \textbf{0.96} & \textbf{0.94} & \textbf{-0.02}
& \textbf{0.95} & \textbf{0.94} & \textbf{-0.01}
& \textbf{0.94} & \textbf{0.93} & \textbf{-0.01} \\
& LPIPS$\downarrow$ 
& \textbf{0.09} & \textbf{0.10} & \textbf{0.01}
& 0.11 & 0.33 & 0.22
& \textbf{0.08} & \textbf{0.09} & \textbf{0.00}
& \textbf{0.08} & \textbf{0.25} & 0.17
& \textbf{0.09} & \textbf{0.19} & 0.09
& \textbf{0.10} & \textbf{0.25} & \textbf{0.16}
& 0.11 & \textbf{0.25} & \textbf{0.14}
& 0.33 & \textbf{0.18} & \textbf{-0.15} \\
\bottomrule
\end{tabular}}
\end{table*}

{\scriptsize
\begin{table*}[t]
\centering
\renewcommand\arraystretch{0.75}
\caption{Rendering and Camera Tracking Results on Replica and TUM Datasets under Different Conditions. “-” indicates reconstruction failure (Of0 in Replica) or missing results due to unavailable settings in the baselines (MonoGS). R0–R2 and Of0–Of4 correspond to room0–2 and office0–4 sequences of Replica (ATE RMSE $\downarrow$ [cm])}
\label{tab:4}
\resizebox{1\textwidth}{!}{
\begin{threeparttable}
\begin{tabular}{ccccccccccc c ccccccc}
\cmidrule[0.4pt]{1-19}
\multirow{2}{*}{Method} & \multirow{2}{*}{Metrics} & \multicolumn{9}{c}{Replica} & & \multicolumn{7}{c}{TUM} \\
\cmidrule[0.3pt]{3-11} \cmidrule[0.3pt]{13-19}
& & Avg. & R0 & R1 & R2 & Of0 & Of1 & Of2 & Of3 & Of4 & & Avg. & fr1/desk & fr1/desk2 & fr1/room & fr2 & fr3 \\
\cmidrule[0.4pt]{1-19}
\multirow{4}{*}{\makecell{SplaTAM\\(clean)}} & ATE/cm↓ & 0.36 & 0.31 & 0.40 & 0.29 & 0.47 & 0.27 & 0.29 & 0.32 & 0.55 & & 5.48 & 3.35 & 6.54 & 11.13 & 1.24 & 5.16 \\
& PSNR↑ & 34.11 & 32.66 & 33.89 & 35.25 & 38.26 & 39.17 & 31.97 & 29.70 & 31.81 & & 22.95 & 22.28 & 22.65 & 21.56 & 25.25 & 23.03 \\
& SSIM↑ & 0.97 & 0.97 & 0.98 & 0.98 & 0.98 & 0.99 & 0.96 & 0.96 & 0.98 & & 0.88 & 0.86 & 0.84 & 0.85 & 0.95 & 0.91 \\
& LPIPS↓ & 0.10 & 0.08 & 0.12 & 0.08 & 0.09 & 0.09 & 0.10 & 0.12 & 0.15 & & 0.19 & 0.23 & 0.25 & 0.21 & 0.09 & 0.17 \\
\cmidrule[0.3pt]{1-19}
\multirow{4}{*}{\makecell{SplaTAM\\(noise)}} & ATE/cm↓ & 0.68 & 0.42 & 1.78 & 0.43 & 4.42 & 0.72 & 0.39 & 0.56 & 0.47 & & 11.69 & 34.4 & 5.93 & 11.4 & 1.52 & 5.20 \\
& PSNR↑ & 24.40 & 24.93 & 23.88 & 24.78 & 20.91 & 25.98 & 24.65 & 22.83 & 23.78 & & 19.56 & 19.48 & 17.06 & 19.15 & 22.04 & 20.07 \\
& SSIM↑ & 0.76 & 0.89 & 0.73 & 0.73 & 0.71 & 0.75 & 0.76 & 0.75 & 0.73 & & 0.75 & 0.77 & 0.65 & 0.72 & 0.86 & 0.75 \\
& LPIPS↓ & 0.58 & 0.42 & 0.52 & 0.61 & 0.70 & 0.69 & 0.60 & 0.59 & 0.50 & & 0.37 & 0.35 & 0.42 & 0.32 & 0.31 & 0.45 \\
\cmidrule[0.3pt]{1-19}
\multirow{4}{*}{\makecell{SplaTAM\\(noise\&low-light)}} & ATE/cm↓ & 6.82 & 0.56 & 2.03 & 2.36 & - & 41.3 & 0.38 & 0.59 & 0.50 & & 12.23 & 34.8 & 6.37 & 12.1 & 2.47 & 5.40 \\
& PSNR↑ & 23.56 & 24.31 & 23.78 & 24.72 & - & 20.78 & 24.20 & 22.97 & 24.15 & & 19.23 & 18.03 & 18.10 & 19.31 & 21.33 & 19.38 \\
& SSIM↑ & 0.73 & 0.81 & 0.72 & 0.75 & - & 0.64 & 0.76 & 0.75 & 0.73 & & 0.72 & 0.68 & 0.63 & 0.73 & 0.84 & 0.72 \\
& LPIPS↓ & 0.59 & 0.41 & 0.53 & 0.61 & - & 0.72 & 0.59 & 0.58 & 0.68 & & 0.43 & 0.43 & 0.44 & 0.34 & 0.33 & 0.53 \\
\cmidrule[0.3pt]{1-19}
\multirow{4}{*}{\makecell{MonoGS\\(clean)}} & ATE/cm↓ & 0.58 & 0.44 & 0.32 & 0.31 & 0.44 & 0.52 & 0.23 & 0.17 & 2.25 & & 1.47 & 1.50 & - & - & 1.44 & 1.49 \\
& PSNR↑ & 34.83 & 36.43 & 37.49 & 39.95 & 42.09 & 36.24 & 36.70 & 36.07 & 37.50 & & 24.43 & 23.63 & - & - & 24.93 & 24.73 \\
& SSIM↑ & 0.96 & 0.95 & 0.96 & 0.97 & 0.97 & 0.98 & 0.96 & 0.96 & 0.96 & & 0.75 & 0.78 & - & - & 0.79 & 0.67 \\
& LPIPS↓ & 0.07 & 0.07 & 0.08 & 0.08 & 0.07 & 0.05 & 0.08 & 0.07 & 0.10 & & 0.25 & 0.32 & - & - & 0.21 & 0.21 \\
\cmidrule[0.3pt]{1-19}
\multirow{4}{*}{\makecell{MonoGS\\(noise)}} & ATE/cm↓ & 1.29 & 0.51 & 0.68 & 0.45 & 1.38 & 2.78 & 0.34 & 0.53 & 0.37 & & 6.32 & 1.53 & - & - & 1.54 & 15.89 \\
& PSNR↑ & 26.94 & 26.36 & 26.69 & 26.45 & 26.79 & 29.12 & 27.13 & 27.02 & 25.82 & & 17.44 & 18.68 & - & - & 15.62 & 18.03 \\
& SSIM↑ & 0.48 & 0.55 & 0.48 & 0.45 & 0.45 & 0.54 & 0.46 & 0.47 & 0.44 & & 0.40 & 0.45 & - & - & 0.41 & 0.34 \\
& LPIPS↓ & 0.63 & 0.55 & 0.66 & 0.68 & 0.82 & 0.65 & 0.64 & 0.52 & 0.74 & & 0.55 & 0.52 & - & - & 0.49 & 0.65 \\
\cmidrule[0.3pt]{1-19}
\multirow{4}{*}{\makecell{MonoGS\\(noise\&low-light)}} & ATE/cm↓ & 1.35 & 0.59 & 0.87 & 0.54 & 1.69 & 2.86 & 0.43 & 0.37 & 3.78 & & 6.36 & 1.55 & - & - & 1.44 & 16.10 \\
& PSNR↑ & 26.93 & 25.57 & 26.62 & 26.98 & 26.18 & 28.01 & 26.74 & 27.24 & 26.31 & & 16.34 & 18.84 & - & - & 13.41 & 16.78 \\
& SSIM↑ & 0.47 & 0.53 & 0.47 & 0.45 & 0.41 & 0.47 & 0.47 & 0.47 & 0.43 & & 0.34 & 0.42 & - & - & 0.32 & 0.29 \\
& LPIPS↓ & 0.65 & 0.52 & 0.66 & 0.72 & 0.84 & 0.75 & 0.64 & 0.59 & 0.74 & & 0.49 & 0.32 & - & - & 0.47 & 0.69 \\
\hline
\end{tabular}
\end{threeparttable}
}
\end{table*}}

\begin{table*}[t]
\scriptsize
\centering
\renewcommand\arraystretch{0.75}
\caption{Rendering and Camera Tracking Results on Replica and TUM Datasets under noisy and low-light Conditions. “-” indicates reconstruction failure (Of0 in Replica) or missing results due to unavailable settings in the baselines (MonoGS). R0–R2 and Of0–Of4 correspond to room0–2 and office0–4 sequences of Replica (ATE RMSE $\downarrow$ [cm])}
\label{tab:5}
\resizebox{1\textwidth}{!}{
\begin{threeparttable}
\begin{tabular}{ccccccccccc c ccccccc}
\cmidrule[0.4pt]{1-19}
\multirow{2}{*}{Method} & \multirow{2}{*}{Metrics} & \multicolumn{9}{c}{Replica} & & \multicolumn{7}{c}{TUM} \\
\cmidrule[0.3pt]{3-11} \cmidrule[0.3pt]{13-19}
& & Avg. & R0 & R1 & R2 & *Of0 & Of1 & Of2 & Of3 & Of4 & & Avg. & fr1/desk & fr1/desk2 & fr1/room & fr2/xyz & fr3/off \\
\cmidrule[0.4pt]{1-19}
\multirow{4}{*}{\makecell{SplaTAM}} & ATE/cm$\downarrow$ & 6.82 & 0.56 & 2.03 & 2.36 & - & 41.3 & 0.38 & 0.59 & 0.50 & & 12.23 & 34.8 & 6.37 & 12.1 & 2.47 & 5.40 \\
& PSNR$\uparrow$ & 23.56 & 24.31 & 23.78 & 24.72 & - & 20.78 & 24.20 & 22.97 & 24.15 & & 19.23 & 18.03 & 18.10 & 19.31 & 21.33 & 19.38 \\
& SSIM$\uparrow$ & 0.73 & 0.81 & 0.72 & 0.75 & - & 0.64 & 0.76 & 0.75 & 0.73 & & 0.72 & 0.68 & 0.63 & 0.73 & 0.84 & 0.72 \\
& LPIPS$\downarrow$ & 0.59 & 0.41 & 0.53 & 0.61 & - & 0.72 & 0.59 & 0.58 & 0.68 & & 0.43 & 0.43 & 0.44 & 0.34 & 0.33 & 0.53 \\
\cmidrule[0.3pt]{1-19}
\multirow{4}{*}{\makecell{MonoGS}} & ATE/cm$\downarrow$ & 1.35 & 0.59 & 0.87 & 0.54 & 1.69 & 2.86 & 0.43 & 0.37 & 3.78 & & 6.36 & 1.55 & - & - & 1.44 & 16.10 \\
& PSNR$\uparrow$ & 26.93 & 25.57 & 26.62 & 26.98 & 26.18 & 28.01 & 26.74 & 27.24 & 26.31 & & 16.34 & 18.84 & - & - & 13.41 & 16.78 \\
& SSIM$\uparrow$ & 0.47 & 0.53 & 0.47 & 0.45 & 0.41 & 0.47 & 0.47 & 0.47 & 0.43 & & 0.34 & 0.42 & - & - & 0.32 & 0.29 \\
& LPIPS$\downarrow$ & 0.65 & 0.52 & 0.66 & 0.72 & 0.84 & 0.75 & 0.64 & 0.59 & 0.74 & & 0.49 & 0.32 & - & - & 0.47 & 0.69 \\
\cmidrule[0.3pt]{1-19}
\multirow{4}{*}{\makecell{Ours}} & ATE/cm$\downarrow$ & \textbf{0.60} & \textbf{0.32} & \textbf{0.76} & \textbf{0.36} & \textbf{1.10} & \textbf{1.30} & \textbf{0.66} & \textbf{0.40} & \textbf{0.43} & & \textbf{2.63} & \textbf{1.52} & \textbf{5.80} & \textbf{11.40} & \textbf{1.34} & \textbf{5.03} \\
& PSNR$\uparrow$ & \textbf{28.05} & \textbf{31.68} & \textbf{28.28} & \textbf{28.62} & \textbf{27.01} & \textbf{27.17} & \textbf{25.76} & \textbf{27.19} & \textbf{27.62} & & \textbf{23.52} & \textbf{23.47} & \textbf{20.86} & \textbf{21.56} & \textbf{26.47} & \textbf{20.63} \\
& SSIM$\uparrow$ & \textbf{0.91} & \textbf{0.97} & \textbf{0.93} & \textbf{0.96} & \textbf{0.81} & \textbf{0.83} & \textbf{0.93} & \textbf{0.92} & \textbf{0.83} & & \textbf{0.90} & \textbf{0.84} & \textbf{0.82} & \textbf{0.82} & \textbf{0.97} & \textbf{0.89} \\
& LPIPS$\downarrow$ & \textbf{0.21} & \textbf{0.08} & \textbf{0.18} & \textbf{0.13} & \textbf{0.36} & \textbf{0.43} & \textbf{0.18} & \textbf{0.19} & \textbf{0.30} & & \textbf{0.16} & \textbf{0.22} & \textbf{0.29} & \textbf{0.28} & \textbf{0.06} & \textbf{0.20} \\
\cmidrule[0.4pt]{1-19}
\end{tabular}
\end{threeparttable}
}
\end{table*}

\subsection{Performance Under Clean Conditions}

We first evaluate tracking accuracy on the clean Replica and TUM datasets. 
Results are summarized in Table~\ref{tab:1} and Table~\ref{tab:2}.  

On Replica dataset, NeRF-based methods including NICE-SLAM~\cite{zhu2022nice}, Co-SLAM~\cite{zhu2023coslam}, Point-SLAM~\cite{zheng2023point}, and Vox-Fusion~\cite{murez2020atlas} achieve ATE values ranging from 0.54 to 1.06 cm, demonstrating their capability for dense mapping but with limited accuracy. 
3DGS-based methods show stronger performance: GS-SLAM~\cite{chen2023gsslam} reaches an average ATE of 0.50 cm, MonoGS~\cite{zimmermann2024monogs} achieves 0.58 cm, and SplaTAM~\cite{hsieh2023splatam} improves further to 0.36 cm. 
Our method obtains the best overall accuracy with an average ATE of 0.24 cm, outperforming the strongest baseline by approximately 50\%. 
Across all individual Replica sequences, our framework consistently ranks first or second, indicating stable performance across diverse scenes.

On TUM dataset, a similar trend is observed. 
Among NeRF-based methods, NICE-SLAM records an average ATE of 15.87 cm, Co-SLAM achieves 8.38 cm, and Point-SLAM reduces error to 8.92 cm, while Vox-Fusion remains at 11.31 cm. 
In contrast, 3DGS-based methods provide superior accuracy: GS-SLAM reports 3.50 cm, MonoGS achieves 1.47 cm on three sequences, and SplaTAM reaches 5.48 cm. 
Our approach again achieves the best performance with an average ATE of 4.79 cm across five sequences, reducing error relative to SplaTAM by more than 12\%. 
In particular, our system attains the lowest ATE values on \textit{fr1/desk}, \textit{fr2/xyz}, and \textit{fr3/off}, demonstrating both accuracy and generalizability to real-world sensor data.

\subsection{Evaluation under Natural Sensor Noise}

To emulate realistic sensor degradation, we construct a natural-noise version of the Replica dataset by injecting physically motivated perturbations into clean frames. 
Following the standard Poisson–Gaussian model of image sensor noise~\cite{foi2008noise,plotz2017benchmark}, photon shot noise is simulated with a Poisson process whose variance depends on the signal intensity, while readout noise is modeled as additive Gaussian disturbance. 
We set the shot noise variance to $3\times 10^{-4}$ $\sim$ $5\times 10^{-4}$ per channel and the readout noise variance to $3\times 10^{-5}$, corresponding to mild noise levels observed in commodity RGB-D sensors. 
Compared with simple Gaussian perturbations, this construction better captures the heteroscedastic and signal-dependent nature of real sensor noise. 
The realism is further confirmed by residual variance measurements: on clean inputs, $\sigma_{R}^{2}$ is 6.84 and 5.92 for \textit{room0} and \textit{room1} respectively, which increase to 15.08 and 14.77 after noise injection (Fig.~\ref{fig:5}). 
These values remain below the strong-noise threshold of $\sigma_{R}^{2} = 30$, showing that the constructed dataset introduces realistic yet not excessively severe perturbations.

Quantitative results are reported in Table~\ref{tab:3}. 
Our method consistently outperforms the baseline SplaTAM across all sequences. 
More importantly, robustness under noise is significantly enhanced: on \textit{office3}, SplaTAM suffers a PSNR drop of 1.48\,dB, while our framework decreases by only 0.55\,dB. 
These results indicate that our system is less sensitive to sensor-induced perturbations, even without invoking the CLIP enhancement module.

Qualitative comparisons in Fig.~\ref{fig:6} further support these findings. 
The top row a1–c1 shows reconstructions from SplaTAM under real sensor noise, which exhibit amplified artifacts and structural distortions. 
In contrast, our method a2–c2 produces cleaner geometry, sharper edges, and reduced noise interference, as highlighted in the red boxes.

\begin{figure}
    \centering
    \includegraphics[width=1\linewidth]{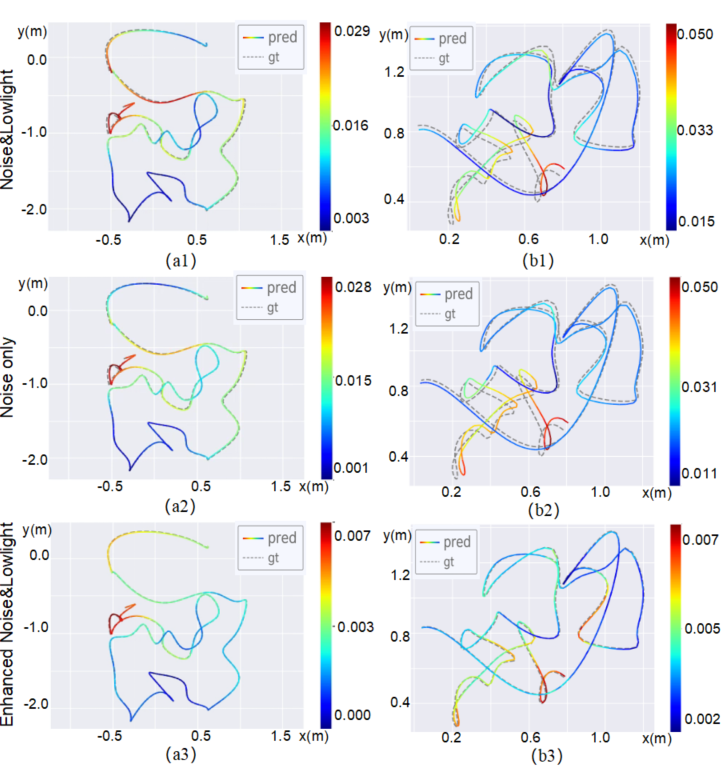}
    \caption{ Comparison of camera motion trajectories under noisy and low-light conditions. 
Rows (a) and (b) correspond to two sequences from the Replica dataset. 
(a1–c1) enhanced noise and low-light, (a2–c2) noise only, and (a3–c3) compounded noise with low-light. Our method achieves trajectories closer to ground-truth under all degraded conditions.}

    \label{fig:8}
\end{figure}

\subsection{Evaluation under Coupled Noise and Low-Light}

To evaluate robustness under compounded degradations, we construct a benchmark by injecting Gaussian noise and applying gamma correction to the Replica and TUM datasets.
Following established practices in computational photography~\cite{hasinoff2014photon,loh2019exdark}, Gaussian noise with a standard deviation of fifteen is employed to simulate sensor perturbations such as thermal noise and dark current. Gamma correction is employed to simulate low-light conditions by decreasing global brightness while maintaining relative contrast.
The correction parameter is set to 1.55 for Replica and 2.25 for TUM. The smaller value for Replica reflects its intrinsically darker pixel intensities, particularly in sequences Of0 and Of1, thereby ensuring that both datasets correspond to human visual perception under low-light conditions.

The quantitative results in Tables~\ref{tab:4} and~\ref{tab:5} demonstrate that compounded degradations severely destabilize current 3DGS-SLAM systems.
On the Replica dataset, the average ATE of SplaTAM increases from 0.36 cm under clean inputs to 6.82 cm under coupled noise and low-light, while SSIM decreases from 0.97 to 0.73.
A similar trend is observed on the TUM dataset, where the ATE rises from 5.48 cm to 12.23 cm.
These results reveal that low-light does not merely introduce additional noise but interacts with photon-limited conditions in a way that amplifies sensor noise and disrupts photometric optimization, leading to a much stronger degradation than noise alone.
By contrast, our method sustains an ATE of 0.60 cm on Replica and 2.63 cm on TUM, achieving a significant reduction in drift relative to the baseline.
In addition, PSNR improves by approximately three to five decibels, and SSIM recovers to values as high as 0.90. Reconstruction performance is shown in Fig.~\ref{fig:7}, and the trajectory comparisons in Fig.~\ref{fig:8} further corroborate these findings, illustrating that our method produces motion estimates that align much more closely with ground-truth even under compounded noisy and low-light conditions.

\begin{figure}
    \centering    \includegraphics[width=1\linewidth]{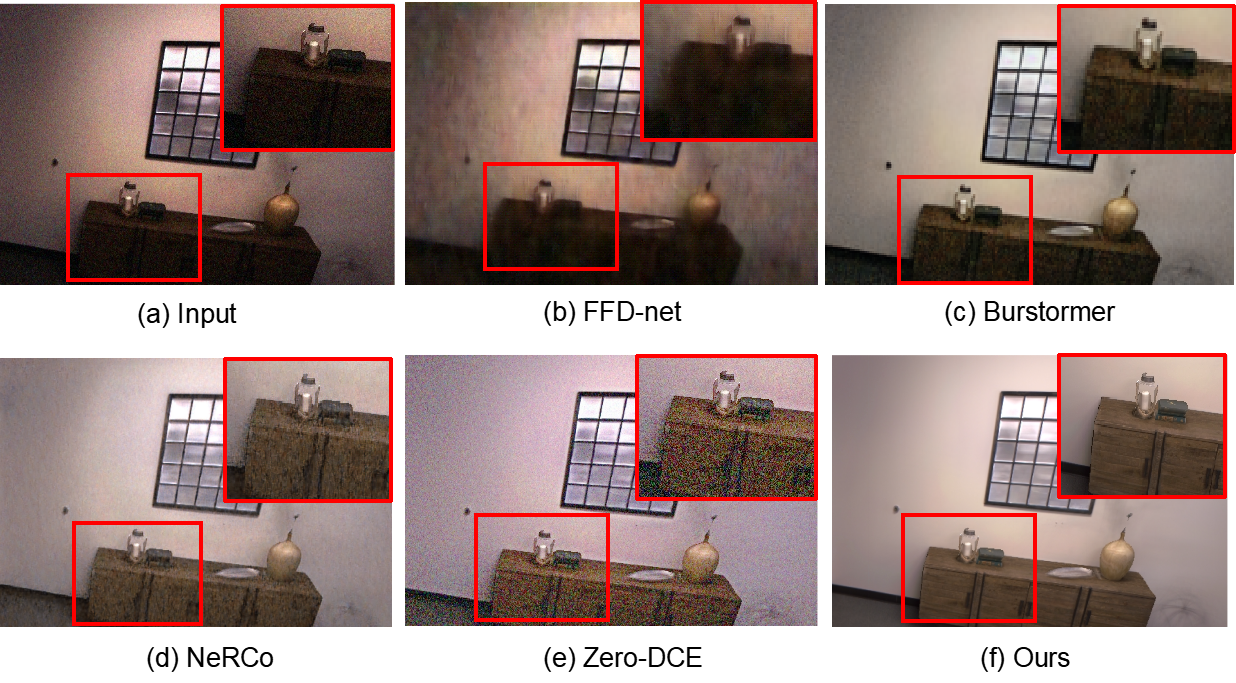}
    \caption{Visual comparison of noise and low-light image enhancement. 
(a) Input degraded image; (b) FFD-net; (c) Burstormer; (d) NeRCo; (e) Zero-DCE; (f) our CLIP-based enhancement. 
Competing methods either over-smooth details or amplify noise, while our approach effectively suppresses noise and restores structural fidelity, producing cleaner and more visually consistent results.}
    \label{fig:9}
\end{figure}

Qualitative comparisons in Fig.~\ref{fig:9} further illustrate the effectiveness of the proposed framework.
FFD-net\cite{zhang2018ffdnet} and Burstormer~\cite{burstormer} are representative denoising networks, while NeRCo~\cite{Nerco} and Zero-DCE~\cite{guo2020zerodce} are low-light enhancement networks that also claim denoising capability.
Although Burstormer is designed to address burst imaging and low-light scenarios, all of these methods either over-smooth structural details or amplify noise under compounded low-light and noise conditions.
In contrast, our CLIP-based enhancement consistently suppresses noise while preserving structural fidelity across diverse sequences.
This generalization ability arises from the intrinsic universality of the CLIP pre-trained model, which enables adaptation to a wide range of degradations without retraining.

\begin{table}[t]
\centering
\large
\caption{Runtime on Replica/R0. 
The additional cost of the CLIP enhancement module is measured per frame. \textcolor{green!50!black}{\textbf{Green}}=best, 
    \textcolor{yellow!50!black}{Yellow}=second, 
    \textcolor{orange!80!black}{Orange}=third}
\label{tab:6}
\renewcommand{\arraystretch}{1.5} 
\resizebox{\linewidth}{!}{
\begin{tabular}{lcccccc}
\cmidrule[0.8pt]{1-7}
Method & \makecell{Tracking\\/Iter.} &\makecell{ Mapping\\/Iter.} & \makecell{Tracking\\/Frame} & \makecell{Mapping\\/Frame} & \makecell{ATE RMSE \\ /[cm]$\downarrow$ }& Gaussian Number \\
\cmidrule[0.8pt]{1-7}
NICE-SLAM~\cite{zhu2022nice} &
30 ms &
166 ms &
1.18 s &
\cellcolor{yellow!25}2.04 s &
0.97 & -- \\
Point-SLAM~\cite{zheng2023point} &
\cellcolor{green!25}19 ms &
\cellcolor{yellow!25}30 ms &
\cellcolor{green!25}0.76 s &
4.50 s &
\cellcolor{orange!25}0.61 & -- \\
SplaTAM~\cite{hsieh2023splatam} &
\cellcolor{orange!25}25 ms &
\cellcolor{green!25}24 ms &
\cellcolor{orange!25}1.00 s &
\cellcolor{green!25}1.44 s &
\cellcolor{yellow!25}0.27 & \cellcolor{yellow!25}5.0\,M \\
Ours &
\cellcolor{yellow!25}22 ms &
\cellcolor{orange!25}34 ms &
\cellcolor{yellow!25}0.89 s &
\cellcolor{orange!25}2.05 s &
\cellcolor{green!25}0.24 & \cellcolor{green!25}1.4\,M \\
\cmidrule[0.8pt]{1-7}
\multicolumn{7}{l}{\textit{+ CLIP enhancement:} ~ 46 ms per frame} \\
\cmidrule[0.8pt]{1-7}
\end{tabular}}
\end{table}

\begin{figure}
    \centering    \includegraphics[width=0.7\linewidth]{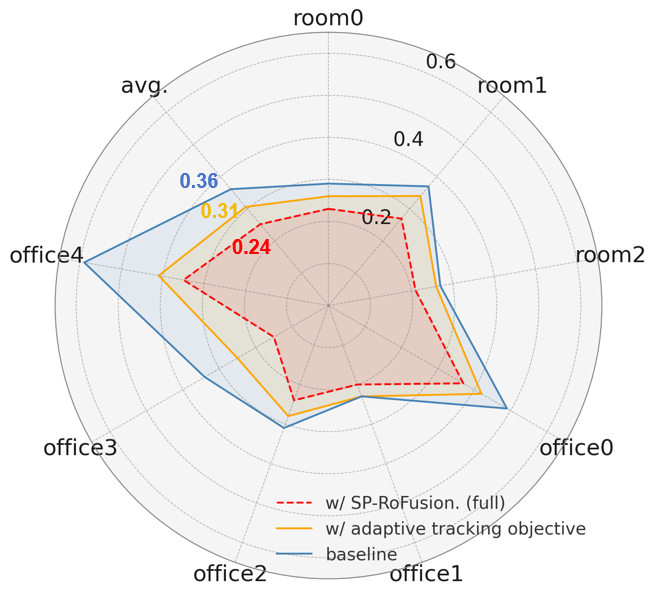}
    \caption{Ablation study on the Replica dataset. 
The radar plot reports ATE (cm, lower is better) across individual sequences and the overall average. 
Baseline results are compared against variants with the adaptive tracking objective and the full SP-RoFusion model. 
Both modules contribute to reducing trajectory error, while the complete system achieves the best overall accuracy.}
    \label{fig:11}
\end{figure}

\subsection{Efficiency Analysis}

\begin{figure}
    \centering    \includegraphics[width=1\linewidth]{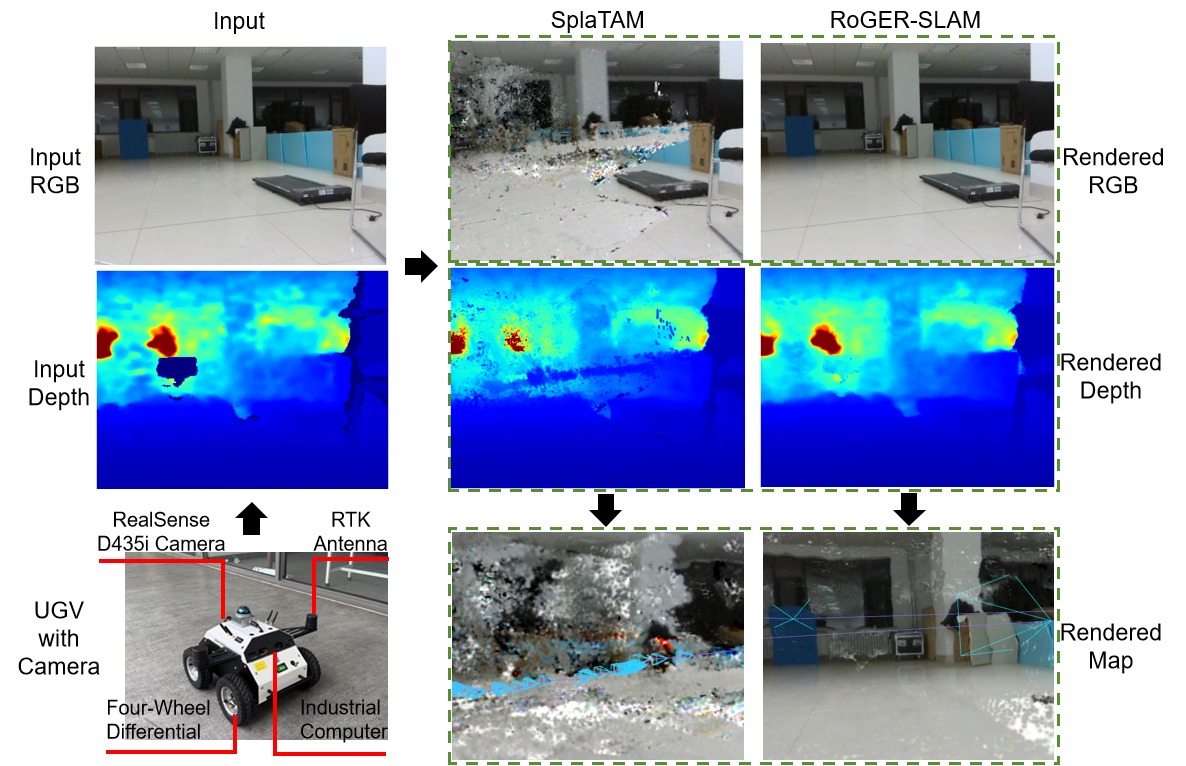}
    \caption{ Real-world experiment on a UGV platform. 
The robot is equipped with an Intel RealSense D435i camera, RTK antenna, and onboard industrial computer (bottom left). 
Given input RGB-D frames (left column), we compare SplaTAM (middle column) with our RoGER-SLAM (right column). 
RoGER-SLAM produces cleaner rendered RGB and depth images, and generates a more complete and geometrically consistent map under real sensor noise.
}
    \label{fig:10}
\end{figure}

We further evaluate the runtime performance of the proposed system on the Replica \textit{room0} sequence, with results summarized in Table~\ref{tab:6}. 
Compared with existing dense SLAM baselines, our framework achieves competitive efficiency, and
the overall accuracy is the best among all methods.

The integration of SP-RoFusion inevitably introduces additional computational overhead, slightly increasing the mapping cost compared with SplaTAM. 
To mitigate this, we adopt a multi-scale gating strategy for densification and an importance-driven pruning mechanism that evaluates each Gaussian using the weighted sum of its alpha opacity and transmittance across multiple views. 
This design effectively suppresses redundant Gaussians and maintains efficiency while preserving mapping fidelity.

In addition, the CLIP enhancement module is selectively activated only under coupled noisy and low-light conditions. 
This conditional triggering avoids unnecessary computation in normal scenarios, adding  46\,ms per frame when invoked. 

\begin{table}[t]
\centering
\Large
\renewcommand{\arraystretch}{1.5}
\caption{Ablation study on Replica (ATE in cm, PSNR in dB). 
Values in parentheses denote improvements relative to the baseline. 
CLIP enhancement is only activated under low-light+noise conditions.}
\label{tab:7}
\resizebox{\linewidth}{!}{
\begin{tabular}{lcccccc}
\cmidrule[0.8pt]{1-7}
& \multicolumn{2}{c}{Clean} & \multicolumn{2}{c}{Natural Noise} & \multicolumn{2}{c}{Noise+Low-light} \\
\cmidrule[0.8pt]{2-7}
Configuration & ATE$\downarrow$ & PSNR$\uparrow$ & ATE$\downarrow$ & PSNR$\uparrow$ & ATE$\downarrow$ & PSNR$\uparrow$ \\
\cmidrule[0.8pt]{1-7}
Baseline (3DGS) & 0.36 & 33.91 & 0.43 & 32.12 & 6.82 & 23.56 \\
+ Adaptive tracking & 0.31 (-0.05) & \textbf{33.93} (+0.02) & 0.38 (-0.05) & 32.14 (+0.02) & 6.81 (-0.01) & 23.48 (-0.08) \\
+ SP-RoFusion & \textbf{0.24} (-0.11) & 33.84 (-0.09) & \textbf{0.28} (-0.15) & \textbf{32.22} (+0.1) & \textbf{6.78} (-0.03) & \textbf{23.97} (+0.41) \\
+ CLIP enhancement & -- & -- & -- & -- & \textbf{0.60} (-6.22) & \textbf{28.05} (+4.49) \\
\cmidrule[0.8pt]{1-7}
\end{tabular}}
\end{table}

\subsection{Ablation Study}

To investigate the contribution of each module, we perform ablation experiments on the Replica dataset under three settings: clean input, natural noise, and coupled noise+low-light, as summarized in Table~\ref{tab:7}. 
The adaptive tracking objective reduces ATE from 0.36\,cm to 0.31\,cm under clean input and from 0.43\,cm to 0.38\,cm under natural noise, and it maintains PSNR at a comparable level. 
SP-RoFusion further improves trajectory accuracy, lowering ATE to 0.24\,cm and 0.28\,cm, respectively, without degrading photometric quality. 
These results indicate that both modules effectively stabilize optimization even before activating enhancement.

The largest improvements are observed under compounded degradations.
When evaluated with coupled noise and low-light, the baseline system exhibits substantial drift with an ATE of 6.82 cm and a PSNR of 23.6 dB.
With the selective activation of the CLIP module, the proposed framework reduces the ATE to 0.60 cm and raises the PSNR to 28.1 dB.
This corresponds to an improvement of 6.22 cm in trajectory accuracy and 4.5 dB in rendering quality.
As illustrated in Fig.~\ref{fig:11}, these improvements are consistent across individual sequences, confirming that adaptive tracking, SP-RoFusion, and CLIP enhancement address complementary failure modes.

\subsection{Real-World Experiments}

To further validate the robustness of our approach in practical scenarios, we conduct real-world experiments on a mobile UGV platform equipped with an Intel RealSense D435i RGB-D camera, an RTK antenna for high-precision ground-truth localization, and an onboard industrial computer for real-time processing. This hardware configuration reflects a typical robotics deployment where sensing is affected by real sensor noise.

The experiments are carried out in an indoor laboratory environment containing a variety of challenging conditions, including textureless walls and floors, reflective surfaces such as glass and metal objects, and occasional dynamic distractors. These factors introduce degradation modes that are not fully captured by synthetic benchmarks, thereby providing a more realistic evaluation of robustness.

Representative results are illustrated in Fig.~\ref{fig:11}, which compares input RGB-D frames, reconstructions from the baseline SplaTAM system and outputs from our RoGER-SLAM framework. The baseline suffers from notable artifacts: fragmented and incomplete geometry, heavy noise accumulation in rendered depth maps, and unstable camera trajectories that cause global misalignment. In contrast, our method produces markedly cleaner RGB and depth reconstructions, preserves planar structures such as floors and walls, and recovers object boundaries with higher fidelity. Furthermore, the rendered maps from our system exhibit more consistent large-scale geometry, demonstrating improved robustness to sensor-induced degradations.

These real-world results confirm that the proposed framework generalizes beyond synthetic datasets and is capable of maintaining high-quality reconstruction under adverse sensing conditions. This highlights the potential of RoGER-SLAM for practical deployment in field robotics, autonomous navigation, and measurement tasks where resilience to noise and low-light degradations is critical.

\section{Conclusion}\label{sec:5}
In this paper, we present RoGER-SLAM, a robust 3DGS-based SLAM system explicitly designed to address the challenges posed by coupled noisy and low-light conditions. Our framework builds upon the inherent low-pass property of Gaussian rendering and proposes three innovations: a structure-preserving robust fusion mechanism that enforces geometric fidelity, an adaptive tracking objective with residual balancing, and a CLIP-based enhancement module selectively activated under compounded degradations.

Extensive experiments on Replica, TUM, and real-world sequences demonstrate that RoGER-SLAM achieves consistently higher trajectory accuracy and reconstruction quality compared with other 3DGS-SLAM systems. The results also provide the first systematic quantitative evidence of how noise and low-light jointly destabilize rendering-driven SLAM, highlighting the importance of explicitly modeling perceptual degradations in future designs. The implementation of RoGER-SLAM will be released as open-source after the review process to facilitate reproducibility and further research.

Looking ahead, future work includes extending the robustness strategies to outdoor environments with dynamic lighting and motion, while incorporating large-scale structural priors for deployment in safety-critical applications. Another promising direction is combining Gaussian-based rendering with inertial or event-based sensing to enhance reliability under extreme conditions.

\section{Acknowledgement}
This work was supported by the State Key Laboratory of Autonomous Intelligent Unmanned Systems and Frontiers Science Center for Intelligent Autonomous Systems, Ministry of Education of China. We are also grateful for the efforts from our colleagues in Sino German Center of Intelligent Systems.

\bibliographystyle{IEEEtran}
\bibliography{reference4}

\begin{IEEEbiography}[{\includegraphics[width=1in,height=1.25in,clip,keepaspectratio]{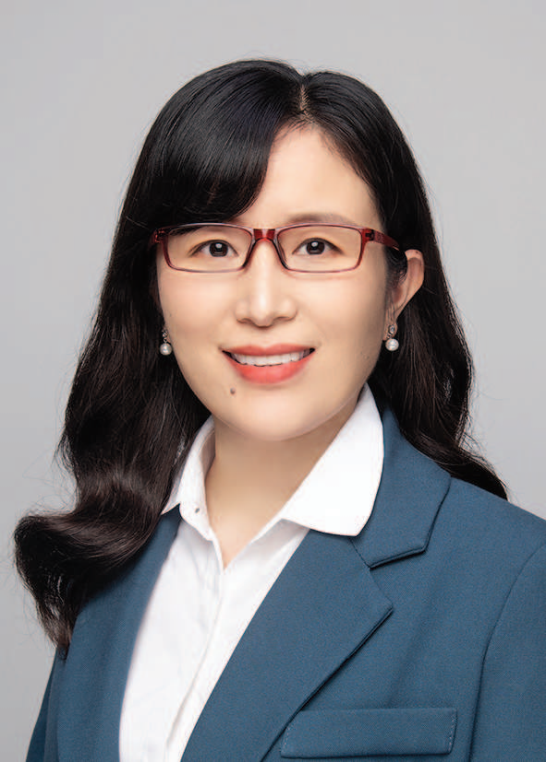}}]{Huilin Yin}
received the Ph.D. degree in control theory and control engineering from Tongji University, China, in 2006. She is currently the chaired professor of T\"{U}EV S\"{U}ED Chair with the Electronic and Information Engineering College, Tongji University. She received the M.S. double-degree from Tongji University and the Technical University of Munich, Germany. Her research interests include the environmental perception of intelligent vehicles and safety for autonomous driving.
\end{IEEEbiography}

\begin{IEEEbiography}[{\includegraphics[width=1in,height=1.25in,clip,keepaspectratio]{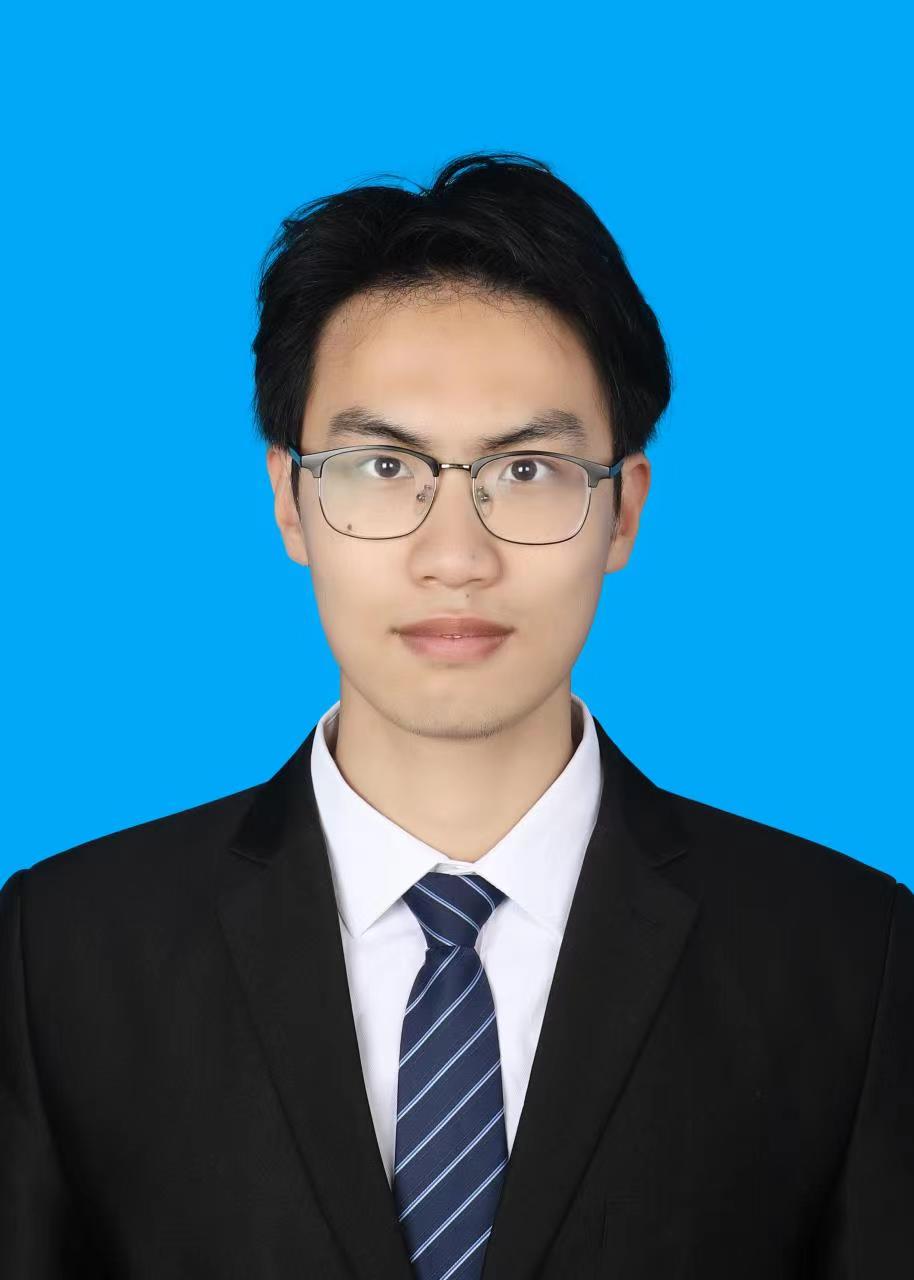}}]{Zhaolin Yang}
Zhaolin Yang received the B.S. degree in electrical engineering and automation from Shaanxi University of Science and Technology, Xi’an, China, in 2023. He is currently pursuing the M.S. degree in electronic information at Tongji University, Shanghai, China, and will join the Technical University of Munich, Munich, Germany, in 2025 as a double-degree master student. His research interests include 3D reconstruction, robotics, deep learning, and SLAM.
\end{IEEEbiography}

\begin{IEEEbiography}
[{\includegraphics[width=1in,height=1.25in,clip,keepaspectratio]{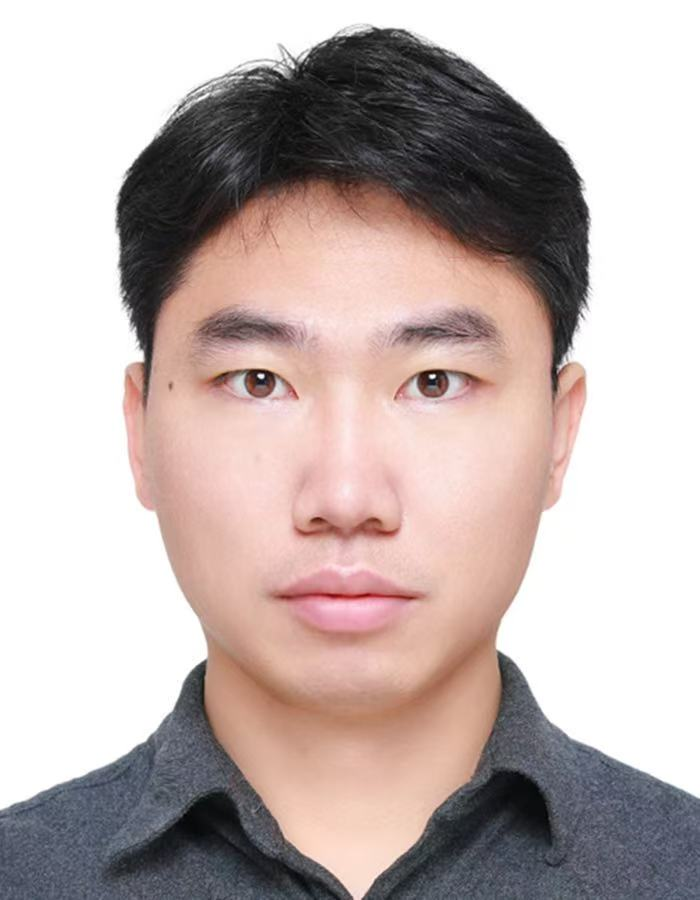}}]{Linchuan Zhang}
received the B.S. degree in intelligent science and technology from Qingdao University in 2019 and the M.S. degree in control engineering from Shandong University, Shandong Province, China, in 2022. He worked as a guest researcher at the Human-Machine Communication Lab of the School of Computer Science, Information and Technology at the Technical University of Munich from February to May 2025. He is currently pursuing the Ph.D. degree in intelligent science and technology at Tongji University, Shanghai, China. His research interests include robotics, deep learning, SLAM, and automation systems.
\end{IEEEbiography}

\begin{IEEEbiography}[{\includegraphics[width=1in,height=1.25in, clip,keepaspectratio]{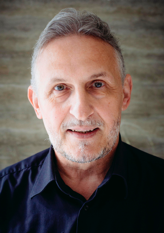}}]{Gerhard Rigoll} 
(Fellow, IEEE) received the Dr.Ing. Habil. degree from the University of Stuttgart, Stuttgart, Germany, in 1991. His Dr. Ing. Habil. thesis was on speech synthesis. In 1993, he was as a Full Professor of computer science at Gerhard Mercator University, Duisburg, Germany. In 2002, he joined the Technical University of Munich (TUM), Munich, Germany, where he is currently heading the Institute for Human-Machine Communication.
\end{IEEEbiography}

\begin{IEEEbiography}[{\includegraphics[width=1in,height=1.25in, clip,keepaspectratio]{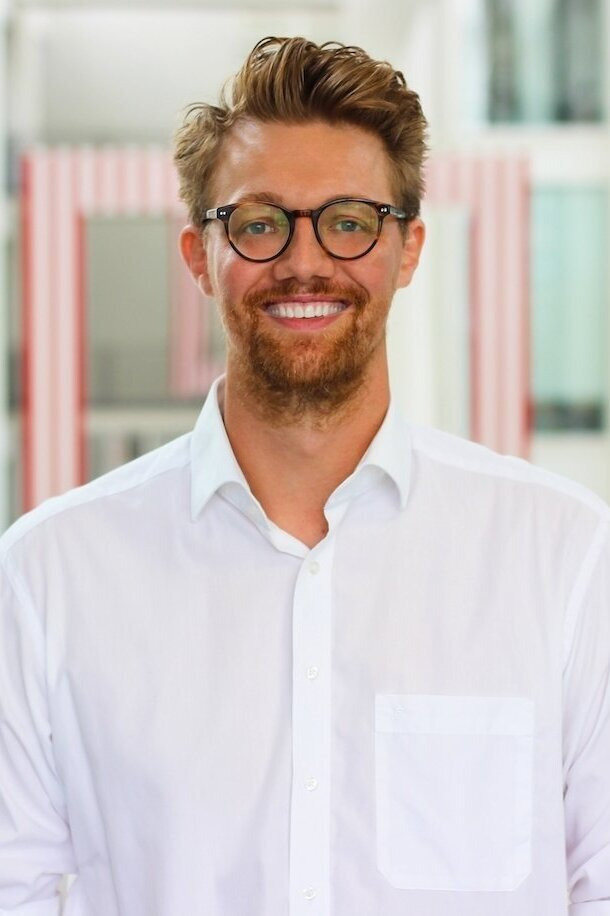}}]{Johannes Betz} 
is an assistant professor in the Department of Mobility Systems Engineering at the Technical University of Munich (TUM). He is one of the founders of the TUM Autonomous Motorsport team. His research focuses on developing adaptive dynamic path planning and control algorithms, decision-making algorithms that work under high uncertainty in multi-agent environments, and validating the algorithms on real-world robotic systems. Johannes earned a B.Eng. (2011) from the University of Applied Science Coburg, an M.Sc. (2012) from the University of Bayreuth, an M.A. (2021) in philosophy from TUM, and a Ph.D. (2019) from TUM. 
\end{IEEEbiography}











\vfill

\end{document}